# Bi-$\ell_0$-$\ell_2$-Norm Regularization for Blind Motion Deblurring

Wen-Ze Shao [a†], Hai-Bo Li [b], Michael Elad [c]

[a] *Department of Computer Science, Technion–Israel Institute of Technology, Haifa 32000, Israel*

*shaowenze1010@163.com*

[b] *School of Computer Science and Communication, KTH Royal Institute of Technology, Stockholm 10044, Sweden*

*haiboli@kth.se*

[c] *Department of Computer Science, Technion–Israel Institute of Technology, Haifa 32000, Israel*

*elad@cs.technion.ac.il*

**Abstract.** In blind motion deblurring, leading methods today tend towards highly non-convex approximations of the $\ell_0$-norm, especially in the image regularization term. In this paper, we propose a simple, effective and fast approach for the estimation of the motion blur-kernel, through a bi-$\ell_0$-$\ell_2$-norm regularization imposed on both the intermediate sharp image and the blur-kernel. Compared with existing methods, the proposed regularization is shown to be more effective and robust, leading to a more accurate motion blur-kernel and a better final restored image. A fast numerical scheme is deployed for alternatingly computing the sharp image and the blur-kernel, by coupling the operator splitting and augmented Lagrangian methods. Experimental results on both a benchmark image dataset and real-world motion blurred images show that the proposed approach is highly competitive with state-of-the- art methods in both deblurring effectiveness and computational efficiency.

## 1. Introduction

Blind motion deconvolution, also known as camera shake deblurring, has been intensively studied since the influential work of Fergus *et al.*[1]. Following the terminology of existing methods [1]-[15], the observed motion-blurred image $y$ is modeled by the spatially invariant convolution, formulated as

$$y = k * x + n, \tag{1}$$

where $x$ is the original image, $k$ is the blur-kernel, $*$ stands for a convolution operator, and $n$ is assumed to be an additive Gaussian noise. The task of blind motion deblurring is generally separated into two independent stages, i.e., estimation of the blur-kernel $k$ and then a non-blind deconvolution of the original image $x$ given the found $k$. The contribution in this paper refers to the first stage, which is the core problem of blind motion deblurring. It is known that this inverse problem is notoriously ill-posed, and therefore appropriate regularization terms or prior assumptions should be imposed in order to achieve reasonable estimates for the sharp image $x$ and the motion blur-kernel $k$. We should emphasize that the by-product estimated image in the first stage is not necessarily a good reconstruction by itself, as indeed observed by state-of-the-art methods [1], [2], [3], [5]-[9], [11]-[15], and its role is primarily to serve the blur-kernel estimation.

[†]Corresponding author. Tel.: +972-584520516.

Most existing motion blur-kernel estimation methods are rooted in the Bayesian framework, with two common kinds of inference principles: Variational Bayes (VB) [1]-[6] and Maximum a Posteriori (MAP) [7]-[15]. The basic idea underlying both principles [1]-[15] is to rely on the Bayes relationship

$$p(\boldsymbol{x}, \boldsymbol{k} \mid \boldsymbol{y}) = \frac{p(\boldsymbol{y} \mid \boldsymbol{x}, \boldsymbol{k})\, p(\boldsymbol{u})\, p(\boldsymbol{k})}{p(\boldsymbol{y})} \propto p(\boldsymbol{y} \mid \boldsymbol{x}, \boldsymbol{k})\, p(\boldsymbol{x})\, p(\boldsymbol{k}). \tag{2}$$

Since the likelihood $p(\boldsymbol{y} \mid \boldsymbol{x}, \boldsymbol{k})$ can be easily formulated due to the Gaussian statistics of the noise $\boldsymbol{n}$, the problem now reduces to the determination of the priors and the posteriori estimation for both the image and the blur-kernel. After a negative log transformation, the MAP estimates of $\boldsymbol{x}$ and $\boldsymbol{k}$ are obtained by computing

$$\min_{\boldsymbol{x}, \boldsymbol{k}}\ \lambda \parallel \boldsymbol{k} * \boldsymbol{x} - \boldsymbol{y} \parallel_2^2 + \alpha_x \mathcal{R}_x(\boldsymbol{x}) + \alpha_k \mathcal{R}_k(\boldsymbol{k}), \tag{3}$$

where $\lambda$, $\alpha_x$, $\alpha_k$ are positive tuning parameters, and $\mathcal{R}_x(\boldsymbol{x})$, $\mathcal{R}_k(\boldsymbol{k})$ are the positive potential functions corresponding to $p(\boldsymbol{x})$ and $p(\boldsymbol{k})$ respectively. In contrast to MAP methods, the VB ones pursue posteriori mean estimates for the image $\boldsymbol{x}$ and the blur-kernel $\boldsymbol{k}$. In Appendix A, we discuss briefly some similarities and differences among existing VB and MAP methods, with emphasis on the choice of the priors for the image and the blur-kernel. Table 1 lists the choice of priors $\mathcal{R}_x(\boldsymbol{x})$ and $\mathcal{R}_k(\boldsymbol{k})$ for the sharp image and the blur-kernel in recent several state-of-the-art MAP methods [10]-[15] and VB methods [5], [6] (referring to the noiseless case).

It is observed that the $\ell_p$-norm-based image prior in [10] (with $p$ set as a non-increasing sequence while iterating), the normalized sparsity-based image prior in [11], the $\ell_{0.3}$-norm-based image prior in [13], the recent approximate $\ell_0$-norm-based image prior in [14] (with the parameter $\varepsilon$ set as a decreasing sequence while iterating), and the re-weighted $\ell_2$-norm-based image prior in [15] are all highly non-convex unnatural sparse priors, attempting to approximate the $\ell_0$-norm via various strategies[1]. As such, they are quite different from the natural image statistics, e.g., [29]-[31], as commonly advocated in the literature in the context of image denoising and non-blind deblurring. As for MAP methods with implicitly unnatural sparse image priors, e.g., [8], [9], their core idea is to estimate the motion blur-kernel from few step-like salient edges in the original image. Those are predicted by suppressing the weak details in flat regions via Gaussian or bilateral smoothing, while enhancing salient edges by shock filtering along with gradient-thresholding operations. In this sense, current successful MAP approaches actually seek an intermediate sharp image with dominant edges as an important clue to motion blur-kernel estimation, rather than a faithful restored image.

---

[1] Other approximate $\ell_0$-norm terms can be envisioned, such as the Gini-Index [18], but as we shall see hereafter, our approach takes a different route.

Table 1. Priors explicitly imposed on the sharp image and the blur-kernel in

state-of-the-art and the proposed methods

| Method | Type | $\mathcal{R}_x(x)$ | $\mathcal{R}_k(k)$ |
|--------|------|--------------------|--------------------|
| [5],[6] | VB | $\sum_m \log |(\nabla \mathbf{x})_m|$ | $\log \| \mathbf{k} \|_2$ |
| [10] | MAP | $\| \nabla \mathbf{x} \|_p^p, \; p : 0.8 \to 0.6 \to 0.4$ | $\| \nabla \mathbf{k} \|_2$ |
| [11] | MAP | $\frac{\| \nabla \mathbf{x} \|_1}{\| \nabla \mathbf{x} \|_2}$ | $\| \mathbf{k} \|_1$ |
| [12] | MAP | $\| \mathbf{F}\mathbf{x} \|_1, \mathbf{F} \triangleq \text{Framelet}$ | $\| \mathbf{k} \|_2^2 + \tau \| \mathbf{F}\mathbf{k} \|_1$ |
| [13] | MAP | $\| \nabla \mathbf{x} \|_{0.3}^{0.3}$ | $\| \mathbf{k} \|_1$ |
| [14] | MAP | $\sum_m \min(1, \frac{|(\nabla \mathbf{x})_m|^2}{\varepsilon^2})$ | $\| \mathbf{k} \|_2^2$ |
| [15] | MAP | $(\nabla \mathbf{x})^T \mathbf{W}(|\nabla \mathbf{x}|) \nabla \mathbf{x}$ | $\| \mathbf{k} \|_2^2 + \varsigma \| \mathbf{k} \|_{0.5}^{0.5}$ |
| Ours [2] | MAP | $c_x^i \cdot (\| \nabla \mathbf{x} \|_0 + \frac{\beta_x}{\alpha_x} \| \nabla \mathbf{x} \|_2^2)$ | $c_k^i \cdot (\| \mathbf{k} \|_0 + \frac{\beta_k}{\alpha_k} \| \mathbf{k} \|_2^2)$ |

In this work we follow this rationale, but aim for pursing a better intermediate sharp image, which naturally leads to more accurate blur-kernel estimation, hence more successful blind deblurring. We propose a simple, fast and effective MAP-based approach for motion blur-kernel estimation, utilizing a bi-$\ell_0$-$\ell_2$-norm regularization imposed on both the sharp image and the blur-kernel[3], as shown in Table 1. While the $\ell_0$- and $\ell_2$-norms have been extensively used in various forms and approximations in earlier blind deblurring work, the regularization we deploy here is different, and as we shall show hereafter, more effective. Our findings suggest that harnessing the proposed framework, the support of the desired motion blur-kernel can be recovered more precisely and robustly. On one hand, the $\ell_0$-$\ell_2$-norm image regularization has greater potential for producing a higher quality sharp image with more accurate salient edges and less staircase artifacts, therefore leading to better blur-kernel estimation. On the other hand, the $\ell_0$-$\ell_2$-norm kernel regularization is able to further improve the estimation accuracy via sparsifying the motion blur-kernel, so as to reduce those possible moderate or strong isolated points as well as weak components in the estimated blur-kernel. Furthermore, this paper applies a continuation strategy to the bi-$\ell_0$-$\ell_2$-norm regularization in order to boost the performance of blind motion deblurring. We formulate the blur-kernel estimation problem as an alternating estimation of a sharp image and a motion blur-kernel. A fast numerical algorithm is proposed for both estimation problems, by coupling the operator splitting and augmented Lagrangian methods, as well as exploiting the fast Fourier transform (FFT). The proposed motion blur-kernel estimation approach does not require any preprocessing operations such as smoothing or edge enhancement, as in earlier work [8], [9].

---

[2] The parameters $c_x, c_k < 1$ are positive continuation factors applied respectively to the proposed $\ell_0$-$\ell_2$-norm regularization on $\mathbf{x}$ and $\mathbf{k}$. With current estimates $\mathbf{x}_i$ and $\mathbf{k}_i$, the quantity $c_x^i$ denotes $c_x$ to the power of $i$ as alternatingly estimating the next estimates $\mathbf{x}_{i+1}$ and $\mathbf{k}_{i+1}$ (See Section 2 and Section 3 for details).

[3] The proposed $\ell_0$-$\ell_2$-norm regularization on $\mathbf{x}$ or $\mathbf{k}$ is somewhat akin to the elastic net regularization [23], which combines the $\ell_2$- and the $\ell_1$-norms in the ridge and LASSO regression methods [24]. However, our interest here is specifically in $\ell_0$ and not $\ell_1$, as it has been demonstrated both theoretically [2], [6], [22] and empirically [11] that a cost function (3) with an $\ell_1$-norm-based image prior naturally leads to a trivial and therefore a useless solution.

This paper provides extensive experiments on both a benchmark image dataset and real-world motion blurred images to validate and analyze the blind deblurring performance of the proposed method. These experiments demonstrate that the proposed approach is highly competitive when compared to state-of-the-art VB and MAP blind motion deblurring methods in both deblurring effectiveness and computational efficiency. We should note that our approach is also found to be robust to the motion blur-kernel size, as well as the parameter settings, to a large degree.

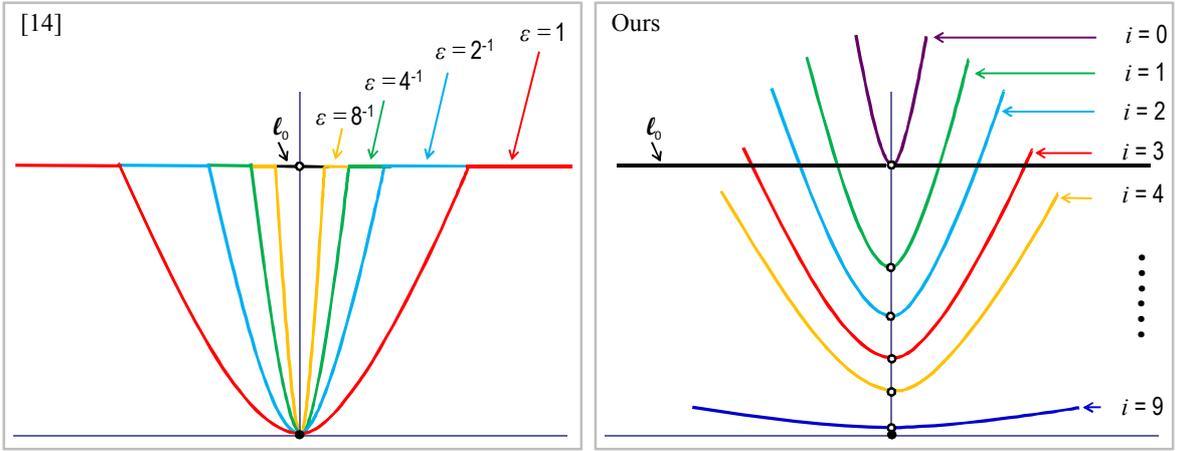

Figure 1. Plots of image priors. [14]: approximate $\ell_0$-norm-based image prior with $\varepsilon$ decreasing from 1 to $8^{-1}$ as iterating; Ours: $\ell_0$-$\ell_2$-norm-based image prior which diminishes though the iterations ($i = 0, 1, ..., I$ -1, and $I$ is set as 10 throughout the paper).

Among the work listed in Table 1, the one by Xu *et al.* [14] with the approximate $\ell_0$-norm image prior and the $\ell_2$-norm kernel prior is similar to the approach proposed in this paper. Both methods attempt to generate an intermediate sharp image for blur-kernel estimation in a strict optimization perspective. However, as observed from the plots of the image priors shown in Figure 1, the working principles of the two methods are fairly distinct. The image prior in [14] approximates the $\ell_0$-norm while iterating for pursing dominant edges as clues for blur-kernel estimation, getting closer and closer to the pure $\ell_0$-norm through the iterations. In contrast, the $\ell_0$-$\ell_2$-norm-based image prior in our scheme is different in several key ways: (i) Our scheme uses the pure $\ell_0$-norm through the iterations, rather than its approximations, which is, however, far more enough; (ii) The augmentation of the $\ell_2$-norm image regularization achieves additional smoothing effect, to a great degree capable of reducing the staircase artifacts ("cartooned" artifacts) in homogenous regions generated by the naive $\ell_0$-norm minimization; and (iii) The continuation strategy adopted in our approach diminishes both the $\ell_0$- and $\ell_2$-norm image regularizations through the iterations. Due to the seeming similarity between the work in [14] and ours, and due to the high-quality performance of [14] (both in speed and output quality)[4], we shall return to discuss the relation between these two works, and provide extensive comparisons between them that demonstrate the superiority of our method.

The paper is organized as follows: Section 2 formulates the motion blur-kernel estimation algorithm using the new bi-$\ell_0$-$\ell_2$-norm regularization. In Section 3, a fast numerical scheme is proposed for the overall problem by coupling the operator splitting strategy and the augmented Lagrangian method. In Section 4, numerous experimental results on Levin *et al.*'s benchmark image dataset and real-world color motion blurred images are provided, accompanied by comparisons with state-of-the-art methods[5]. Section 5 concludes this paper.

## 2. Blind Motion Deblurring Using Bi-$\ell_0$-$\ell_2$-norm Regularization

Intuitively, the accuracy of motion blur-kernel estimation relies heavily on the quality of the sharp image that is reconstructed along with the kernel. It has been shown in [2], [6], [11], [22] that the commonly used natural image statistics, e.g., $\ell_p$-norm-based super-Gaussian prior ($1 \geq p \gg 0$) [29], generally fails to recover the true support of a motion blur kernel. In contrast, the unnatural $\ell_0$-norm-approximating priors (explicitly or implicitly) [5], [6], [8], [9], [11], [13]-[15] are consistently found to perform more effectively, roughly implying that the desired sharp image used in the motion blur-kernel estimation stage should be different from the original image, by putting more emphasis on salient edges while sacrificing weak content.

In this paper, instead of struggling with an approximation to the naive $\ell_0$-norm-based image prior, as in other methods, or directly making use of it, we work directly with a pure $\ell_0$-norm. We formulate the blind motion deblurring problem with a bi-$\ell_0$-$\ell_2$-norm regularization imposed on both the sharp image and the motion blur-kernel. Similar to (3), a cost function based on the new prior is given as follows

$$\min_{\mathbf{x}, \mathbf{k}} \ \lambda \parallel \boldsymbol{k} * \boldsymbol{x} - \boldsymbol{y} \parallel_2^2 + \mathscr{R}_0(\mathbf{x}, \mathbf{k}), \tag{4}$$

where $\mathbf{k}$ is the vectorized representation of $\boldsymbol{k}$ and $\mathscr{R}_0(\mathbf{x}, \mathbf{k})$ is the bi-$\ell_0$-$\ell_2$-norm regularization defined as

$$\mathscr{R}_0(\mathbf{x}, \mathbf{k}) = \alpha_x (\parallel \nabla \mathbf{x} \parallel_0 + \tfrac{\beta_x}{\alpha_x} \parallel \nabla \mathbf{x} \parallel_2^2) + \alpha_k (\parallel \mathbf{k} \parallel_0 + \tfrac{\beta_k}{\alpha_k} \parallel \mathbf{k} \parallel_2^2), \tag{5}$$

---

[4] We should note that when we refer to [14] later in the results section, we actually consider two versions of their work - the one reported in [14], and a combination of [14] and [9] that the authors released later on, due to its better performance. Both versions were taken from the authors' webpage: http://www.cse.cuhk.edu.hk/leojia/deblurring.htm.

[5] Upon publication of this paper, we intend to release a MATLAB software package reproducing the complete set of experiments reported here..

where $\beta_x, \beta_k$ are positive tuning parameters. In this equation, the first two terms correspond to the $\ell_0$-$\ell_2$-norm-based image regularization, and the second two correspond to a similar regularization that serves the motion blur-kernel. The rationale underlying the first part is the desire to get a recovered image with the dominant edges from the original image, which govern the main blurring effect, while also to force smoothness along prominent edges and inside homogenous regions. Such a sharp image is more reliable for recovering the true support of the desired motion blur-kernel than alternative images with unpleasant staircase artifacts. As for the $\ell_0$-$\ell_2$-norm regularization on the blur-kernel, it is rooted in the natural sparseness property of typical motion blur-kernels. This prior leads to an improved estimation precision via sparsifying the motion blur-kernel; the $\ell_0$-norm reduces those possible moderate or strong isolated points in the blur-kernel, and the $\ell_2$-norm part suppresses the weak components just as practiced in [8], [9], [14].

An inherent problem to Equation (5) is the tiresome choice of appropriate regularization parameters. Take the $\ell_0$-$\ell_2$-norm-based image regularization for example. If $\alpha_x, \beta_x$ are set too small throughout the iterations, the regularization effect would be so minor that the estimated image would be too blurred, thus leading to poor quality estimated kernels. On the contrary, if $\alpha_x, \beta_x$ are set too large, the intermediate sharp image will become too "cartooned", which generally has fairly less accurate edge structures accompanied by unpleasant staircase artifacts in the homogeneous areas, thus degrading the kernel estimation precision. To alleviate this problem, a continuation strategy is applied to the bi-$\ell_0$-$\ell_2$-norm regularization (5) so as to reach a compromise. More specifically, assume current estimates of the image and the kernel are $\mathbf{x}_i$ and $\mathbf{k}_i$. The next estimates of $\mathbf{x}_{i+1}$, $\mathbf{k}_{i+1}$ are obtained by solving a modified minimization problem of (4)

$$(\mathbf{x}_{i+1}, \mathbf{k}_{i+1}) = \arg\min_{\mathbf{x},\mathbf{k}} \lambda \parallel \mathbf{k} * \mathbf{x} - \mathbf{y} \parallel_2^2 + \mathscr{R}_1^i(\mathbf{x}, \mathbf{k}), \qquad (6)$$

where $\mathscr{R}_1^i$ is given by

$$\mathscr{R}_1^i(\mathbf{x}, \mathbf{k}) = c_x^i \cdot \alpha_x (\parallel \nabla \mathbf{x} \parallel_0 + \tfrac{\beta_x}{\alpha_x} \parallel \nabla \mathbf{x} \parallel_2^2) + c_k^i \cdot \alpha_k (\parallel \mathbf{k} \parallel_0 + \tfrac{\beta_k}{\alpha_k} \parallel \mathbf{k} \parallel_2^2), \qquad (7)$$

and $c_x, c_k < 1$ are positive continuation factors which are fixed as $2/3$ and $4/5$, respectively, for all the experiments in this paper. With this continuation strategy, the regularization effect is diminishing as we iterate, which leads to more and more accurate salient edges in a progressive manner, and is to be quite beneficial for improving the blur-kernel estimation precision. Note that the continuation strategy is also applied to the $\ell_p$-norm-based image prior in [10], however its continuation factor has to be adjusted for each blind deblurring problem. In addition, the blur-kernel size is set differently for different blurring levels, and as claimed by the authors, it is chosen to be slightly larger than the size of the actual blur. In these respects, our method is more robust and flexible, as we indeed demonstrate in Section 4. Although the optimization problem (6) is highly non-smooth and non-convex, a fast numerical scheme is derived in Section 3, via coupling the operator splitting and the augmented Lagrangian (OSAL) methods.

In spite of numerous work in the past decade, the question: "what is a good prior for blind motion deblurring" remains an open problem. The proposed bi-$\ell_0$-$\ell_2$-norm regularization is mathematically a simple combination of the $\ell_0$- and $\ell_2$-norms, and yet, it is highly effective. We should declare it is not that trivial as it seems to be. In Section 4, numerous experimental results demonstrate that the new regularization term is indeed a better prior compared with the previous ones. To obtain an intuitive understanding of the benefit of the bi-$\ell_0$-$\ell_2$-norm regularization, another two regularization terms with the naive $\ell_0$-norm-based image prior are also considered in this paper, i.e.,

$$\mathscr{R}_2^i(\mathbf{x}, \mathbf{k}) = c_x^i \cdot \alpha_x \parallel \nabla \mathbf{x} \parallel_0 + c_k^i \cdot \alpha_k (\parallel \mathbf{k} \parallel_0 + \tfrac{\beta_k}{\alpha_k} \parallel \mathbf{k} \parallel_2^2), \tag{8}$$

$$\mathscr{R}_3^i(\mathbf{x}, \mathbf{k}) = c_x^i \cdot \alpha_x \parallel \nabla \mathbf{x} \parallel_0 + c_k^i \cdot \beta_k \parallel \mathbf{k} \parallel_2^2, \tag{9}$$

which are degenerated versions of Equation (7) and demonstrated to be inferior to it in Section 4.

## 3. Fast Optimization

### 3.1 Alternating minimization for bi-$\ell_0$-$\ell_2$-regularized blind motion deblurring

Our practical implementation minimizes the cost function in Equation (6) by alternating estimations of the sharp image and the motion blur-kernel. Therefore, this paper addresses the motion blur-kernel estimation as the following alternating $\ell_0$-$\ell_2$-regularized least squares problems with respect to $\mathbf{x}$ and $\mathbf{k}$. First, we estimate the sharp image given the blur-kernel $\mathbf{k}_i$,

$$\mathbf{x}_{i+1} = \arg\min_{\mathbf{u}} \ \lambda \parallel \mathbf{K}_i \mathbf{x} - \mathbf{y} \parallel_2^2 + c_x^i \cdot \alpha_x (\parallel \nabla \mathbf{x} \parallel_0 + \tfrac{\beta_x}{\alpha_x} \parallel \nabla \mathbf{x} \parallel_2^2), \tag{10}$$

where $\mathbf{K}_i \in \mathbf{R}^{M \times M}$ is the BCCB (block-circulant with circulant blocks)[6] convolution matrix corresponding to $\mathbf{k}_i$, $M$ is the number of image pixels, and $\mathbf{y}$ is the vectorized representation of $y$. Turning to the estimate of the kernel given the sharp image $\mathbf{x}_{i+1}$, our empirical tests suggest that this task is better performed when done in the image derivative domain (a similar statement is also made in [14]). Thus,

$$\mathbf{k}_{i+1} = \arg\min_{\mathbf{k}} \lambda \sum_{d \in \Lambda} \parallel (\mathbf{X}_{i+1})_d \mathbf{k} - \mathbf{y}_d \parallel_2^2 + c_k^i \cdot \alpha_k (\parallel \mathbf{k} \parallel_0 + \tfrac{\beta_k}{\alpha_k} \parallel \mathbf{k} \parallel_2^2), \tag{11}$$

where $d \in \Lambda \triangleq \{h, v\}$, $\mathbf{y}_d = \nabla_d \mathbf{y}$, $(\mathbf{x}_{i+1})_d = \nabla_d \mathbf{x}_{i+1}$, $(\mathbf{X}_{i+1})_d$ represents the convolution matrix corresponding to the image gradient $(\mathbf{x}_{i+1})_d$. According to [26], [25], it is known that both the problems posed in (10) and (11) are NP-hard in general. One more point to be noted is that the motion blur-kernel should be non-negative as well as normalized, and therefore the output estimated blur-kernel is projected onto the constraint set $\mathcal{C} = \{\mathbf{k} \geq 0, \ \parallel \mathbf{k} \parallel_1 = 1\}$.

---

[6] The image boundaries are smoothed so as to approximate the circular boundary condition.

---

**Algorithm 1:** Alternating Minimization for Bi-$\ell_0$-$\ell_2$-regularized Blind Motion Deblurring

---

1: **Input :** blurred image $\mathbf{y}$, regularization parameters $\lambda$, $\alpha_x$, $\beta_x$, $\alpha_k$, $\beta_k$, outer iteration
    number $I$, inner iteration numbers $L$, $J$, and continuation factors $c_x$, $c_k$.

2: **Initialization :** $\mathbf{x}_0$, $\mathbf{k}_0$, $i = 0$.

3: **While** $i < I$ do
- Update $\mathbf{x}_{i+1}$ by solving (10) based on **Algorithm 2**.
- Update $\mathbf{k}_{i+1}$ by solving (11) based on **Algorithm 3**.
- Project $\mathbf{k}_{i+1}$ onto the constraint set $\mathcal{C}$.
- Update the parameters based on the continuation factors $c_x$, $c_k$.
- Update $i = i+1$.

4: **End**

5: **Output :** $\mathbf{x}_I$, $\mathbf{k}_I$.

---

The alternating minimization framework for motion blur-kernel estimation requires no extra pre-processing steps such as image smoothing or edge enhancement, which is quite different from other MAP methods [8], [9], [27]. We propose a fast numerical scheme that approximates the required solutions for (10) and (11), by coupling the operator splitting and the augmented Lagrangian (OSAL) methods for both (10) and (11), in the similar spirit to [19], [20]. The pseudo-code of the overall numerical scheme is presented as **Algorithm 1**.

*3.2 OSAL-based $\ell_0$-$\ell_2$-minimization for estimating the sharp image and the motion blur-kernel*

We turn to the OSAL method, used to derive a fast numerical scheme for both (10) and (11). Firstly, apply operator splitting to (10), getting an equivalent constrained minimization problem

$$(\mathbf{w}_{i+1}, \mathbf{x}_{i+1}) = \arg\min_{\mathbf{w}, \mathbf{x}} \frac{\lambda}{c_x^i} \| \mathbf{K}_i \mathbf{x} - \mathbf{y} \|_2^2 + \alpha_x (\| \mathbf{w} \|_0 + \frac{\beta_x}{\alpha_x} \| \nabla \mathbf{x} \|_2^2) \quad \text{s.t.} \quad \mathbf{w} = \nabla \mathbf{x}. \tag{12}$$

Secondly, based on the augmented Lagrangian method, $\mathbf{w}_{i+1}$ and $\mathbf{x}_{i+1}$ can be iteratively estimated by the following unconstrained minimization problem

$$(\mathbf{w}_i^{l+1}, \mathbf{x}_i^{l+1}) = \arg\min_{\mathbf{w}, \mathbf{x}} \frac{\lambda}{c_x^i} \| \mathbf{K}_i \mathbf{x} - \mathbf{y} \|_2^2 + \alpha_x (\| \mathbf{w} \|_0 + \frac{\beta_x}{\alpha_x} \| \nabla \mathbf{x} \|_2^2) + \boldsymbol{\mu}_x^{l*}(\nabla \mathbf{x} - \mathbf{w}) + \frac{\gamma_x}{2} \| \nabla \mathbf{x} - \mathbf{w} \|_2^2, \tag{13}$$

where $0 \leq l \leq L-1$. In the above equation, $\gamma_x$ is the augmented Lagrangian penalty parameter. The Lagrange multiplier, $\boldsymbol{\mu}_x^l$, for the constraint $\mathbf{w} = \nabla \mathbf{x}$ is updated according to the rule

$$\boldsymbol{\mu}_x^{l+1} = \boldsymbol{\mu}_x^l + \gamma_x (\nabla \mathbf{x}_i^{l+1} - \mathbf{w}_i^{l+1}). \tag{14}$$

In principle, the continuation strategy can be also applied to the penalty parameter $\gamma_x$, i.e., $\gamma_x^{l+1} \leftarrow \rho_x \times \gamma_x^l$, with a small initialization $\gamma_x^0 > 0$ and $\rho_x > 1$. However, it is empirically found in this work that a fixed large $\gamma_x$ equal to 100 works well in all the experiments. After some straightforward manipulations, $\mathbf{w}_l^{l+1}$, $\mathbf{x}_i^{l+1}$ can be easily computed from (13) and given as

$$\mathbf{w}_l^{l+1} = \Theta_{\text{Hard}}\left(\nabla \mathbf{x}_i^l + \frac{1}{\gamma_x}\boldsymbol{\mu}_x^l, \ (\frac{2\alpha_x}{\gamma_x})^{\frac{1}{2}}\right), \tag{15}$$

$$\mathbf{x}_i^{l+1} = \left(\frac{\lambda}{c_x^*}\mathbf{K}_i^*\mathbf{K}_i + (\beta_x + \frac{\gamma_x}{2})\nabla^*\nabla\right)^{-1}\left(\frac{\lambda}{c_x^*}\mathbf{K}_i^*\mathbf{y} + \frac{\gamma_x}{2}\nabla^*(\mathbf{w}_l^l - \frac{1}{\gamma_x}\boldsymbol{\mu}_x^l)\right), \tag{16}$$

where $\mathbf{K}_i^*$ is the conjugate transpose of $\mathbf{K}_i$, $\mathbf{x}_0^0, \mathbf{w}_0^0, \boldsymbol{\mu}_x^0$ are the initializations, and $\Theta_{\text{Hard}}$ is the hard- thresholding operator defined as $\Theta_{\text{Hard}}(a,b) = a \cdot (|a| \geq b)$. Then, the minimizers of (12) can be obtained as $\mathbf{w}_{i+1} = \mathbf{w}_l^L$, $\mathbf{x}_{i+1} = \mathbf{x}_i^L$. Equation (15) is computationally very simple to implement because of its pixel-by-pixel processing. Also, in this work a circular convolution is assumed for the observation model (1), and hence (16) can be also computed very efficiently using FFT.

---

**Algorithm 2** OSAL-based $\boldsymbol{\ell}_0$-$\boldsymbol{\ell}_2$-minimization for the Sharp Image Estimation

---

1: **Input :** Motion blur-kernel $\mathbf{k}_i$, penalty parameter $\gamma_x$.

2: **Initialization :** $\mathbf{x}_0^0 = \mathbf{x}_0, \mathbf{w}_0^0, \boldsymbol{\mu}_x^0, l = 0$.

3: **While** $l < L$ do

   - Update $\mathbf{w}_l^{l+1}$ by computing (15).
   - Update $\mathbf{x}_i^{l+1}$ by computing (16) based on FFT.
   - Update $\boldsymbol{\mu}_x^{l+1}$ by computing (14).

4: **End**

5: **Output :** $\mathbf{x}_{i+1} = \mathbf{x}_i^L$.

---

To summarize, the OSAL-based $\boldsymbol{\ell}_0$-$\boldsymbol{\ell}_2$-minimization for the sharp image estimation amounts to iterative computations of (14)-(16). The pseudo-code of the numerical scheme is presented as **Algorithm 2**. We note that a different numerical scheme from the one proposed here is used in [26] and [25] to solve their specific inverse Potts problem with affirmative convergence analysis. Actually, provided that $\gamma_x$ goes to infinity, a similar analysis can be made for **Algorithm 2** by borrowing the core ideas in [26], [25].

The OSAL method is also used to handle the problem posed in (11). Due to the close similarity between the tasks posed by the minimization functionals (10) and (11), we turn directly to the pseudo-code presented as **Algorithm 3**. Similar to $\mathbf{w}_l^{l+1}$ in (15), the computation of $\mathbf{g}_l^{l+1}$ corresponds to a simple pixel-by-pixel thresholding operation; and in the same manner as $\mathbf{x}_i^{l+1}$ in (16), $\mathbf{k}_i^{l+1}$ is efficiently computed using FFT with the circular convolution assumption. Additionally, the augmented Lagrangian penalty parameter $\gamma_k$ is kept fixed to a large value $1 \times 10^6$ in all the experiments.

---

**Algorithm 3:** OSAL-based $\ell_0$-$\ell_2$-minimization for Motion Blur-kernel Estimation

1 : **Input :** sharp image $\mathbf{x}_{i+1}$, penalty parameter $\gamma_k$.

2 : **Initialization :** $\mathbf{k}_0^0 = \mathbf{k}_0$, $\mathbf{g}_0^0$, $\boldsymbol{\mu}_k^0$, $j = 0$.

3 : **While** $j < J$ **do**

- Update $\mathbf{g}_i^{j+1}$ by computing $\mathbf{g}_i^{j+1} = \Theta_{\text{Hard}} \left( \mathbf{k}_i^j + \frac{1}{\gamma_k} \boldsymbol{\mu}_k^j, \ (\frac{2\alpha_k}{\gamma_k})^{\frac{1}{2}} \right)$.

- Update $\mathbf{k}_i^{j+1}$ based on FFT by computing

$$\mathbf{k}_i^{j+1} = \left( \frac{\lambda}{c_k^2} \textstyle\sum_{d \in \Lambda} (\mathbf{X}_{i+1})_d^* (\mathbf{X}_{i+1})_d + (\beta_k + \frac{\gamma_k}{2}) \mathbf{I} \right)^{-1} \left( \frac{\lambda}{c_k^2} \textstyle\sum_{d \in \Lambda} (\mathbf{X}_{i+1})_d^* \mathbf{y}_d + \frac{\gamma_k}{2} (\mathbf{g}_i^j - \frac{1}{\gamma_k} \boldsymbol{\mu}_k^j) \right).$$

- Update $\boldsymbol{\mu}_k^{j+1}$ by computing $\boldsymbol{\mu}_k^{j+1} = \boldsymbol{\mu}_k^j + \gamma_k (\mathbf{k}_i^{j+1} - \mathbf{g}_i^{j+1})$.

4 : **End**

5 : **Output :** $\mathbf{k}_{i+1} = \mathbf{k}_i^j$.

---

*3.3 Other Implementation Details*

In order to account for the large-scale motion blur-kernel estimation as well as to further reduce the risk of getting stuck in a poor local minimum, a multi-scale ($S$ scales) version of **Algorithm 1** is actually used, similar to all top-performing VB [1]-[6] and MAP [7]-[15] methods. The pseudo-code of the multi-scale implementation of **Algorithm 1** is summarized as **Algorithm 4** ($S = 4$). In each scale $s$, the input blurred image $\mathbf{y}_s$ is the 2 times down-sampled blurred image from the original blurred image $\mathbf{y}$ (in the finest scale the input is the original blurred image $\mathbf{y}$ itself), $\mathbf{x}_0^0$ is simply set as a zero image, and $\mathbf{k}_0^0$ is set as the up-sampled blur-kernel from the coarser level (in the coarsest scale $\mathbf{k}_0^0$ is set as a Dirac pulse). As for $\mathbf{w}_0^0, \boldsymbol{\mu}_x^0, \mathbf{g}_0^0, \boldsymbol{\mu}_k^0$, they are also set as zeros. The outer iteration number $I$ and the inner iteration numbers $L$ and $J$ are all set as 10. As for the parameters $\alpha_x$, $\beta_x$, $\alpha_k$, $\beta_k$, and $\lambda$, they are fixed to

$$\alpha_x = 0.25, \ \beta_x = 5, \ \alpha_k = 0.25, \ \beta_k = 5, \ \lambda = 100$$

across all the experiments reported in the present paper. Additionally, the non-blind deblurring algorithm in [16] is used throughout the paper, which is based on the hyper-Laplacian image prior.

---

**Algorithm 4:** Multi-scale Implementation of **Algorithm 1** for Blind Motion Deblurring

1 : **Input :** Scale number $S$, blurred image $\mathbf{y}$, downsampled images $\{\mathbf{y}_s\}$ in coarser scales $s < S$,
   $\mathbf{y}_S = \mathbf{y}$, parameters $\alpha_x, \beta_x, \gamma_x, c_x, \alpha_k, \beta_k, \gamma_k, c_k, \lambda, I, L, J$.

2 : **Initialization :** $s = 1$, $\mathbf{x}_0^0 = \mathbf{0}$, $\mathbf{k}_0^0 = $ Dirac pulse.

3 : **While** $s \leq S$ **do**
   - Estimate the motion blur-kernel for the $s$th scale $\mathbf{k}_s \triangleq \mathbf{k}_I$ using **Algorithm 1**.
   - Initialize $\mathbf{k}_0^0$ by upsampling $\mathbf{k}_s$ with projection onto the set $\mathcal{C}$ for the $(s+1)$th scale.
   - Initialize $\mathbf{x}_0^0$ by $\mathbf{y}_s$ for the $(s+1)$th scale.

4 : **End**

5 : **Output :** $\mathbf{k}_S$.

6 : **Deconvolution :** Estimate the deblurred image $\mathbf{x}$ using the non-blind deblurring method [16].

---

## 4. Experimental Results

### 4.1 Experiments on Levin et al.'s benchmark dataset

In this subsection, the proposed approach is tested on the benchmark image dataset proposed by Levin *et al.* in [2], downloaded from the author's homepage[7]. The dataset contains 32 real motion blurred images generated from 4 natural images of size 255×255 and 8 different motion blur-kernels of sizes ranging from 13×13 to 27×27 estimated by recording the trace of focal reference points on the boundaries of the original images [2], [6]. Accompanying the benchmark dataset, the estimated blur-kernels corresponding to [1], [3], [8] are also provided for ease of comparison. The 4 images and 8 motion blur-kernels are shown in Figure 2. The SSD metric (Sum of Squared Difference) defined in [2] is used to conduct evaluations on all the methods, quantifying the error between the estimated and the original images. As suggested by state-of-the-art methods, e.g., [2], [3], [5], [6], [7], [11], [13], [14], the SSD error ratio between the images deconvolved respectively with the estimated blur-kernel (its size is set the same as the true one) and the ground truth blur-kernel is used as the final evaluation measure. This way, we take into account the fact that a harder blur-kernel gives a larger deblurring error even if the ground truth blur-kernel is known, since the corresponding non-blind deconvolution problem is also harder.

The first experiment we introduce compares blind motion deblurring performance using the proposed bi-$\ell_0$-$\ell_2$ regularization (7) versus its two degenerated versions (8) and (9). The corresponding deblurring algorithms are denoted, respectively, as **Algorithm 4**-(7), **Algorithm 4**-(8), and **Algorithm 4**-(9) in the following text. For fairness, the involved parameters in **Algorithm 4**-(8) and **Algorithm 4**-(9) are tuned and also fixed across the 32 images to achieve the "best" blind deblurring performance. Figure 3 shows the cumulative histogram of the SSD deconvolution error ratios across 32 test images for each algorithm. Following convention of earlier work, the *r*'th bin in the figure counts the percentage of the motion blurred images in the dataset achieving error ratio below *r* [2]. For instance, the bar in Figure 3 corresponding to bin 3 indicates the percentage of test images with SSD error ratios below 3. For each bin, the higher the bar, the better the deblurring performance. As pointed out by Levin *et al.* [2], deblurred images are visually plausible in general if their SSD error ratios are below 3, and in this case the blind motion deblurring is considered to be successful.

---

[7]www.wisdom.weizmann.ac.il/~levina/papers/LevinEtalCVPR2011Code.zip.

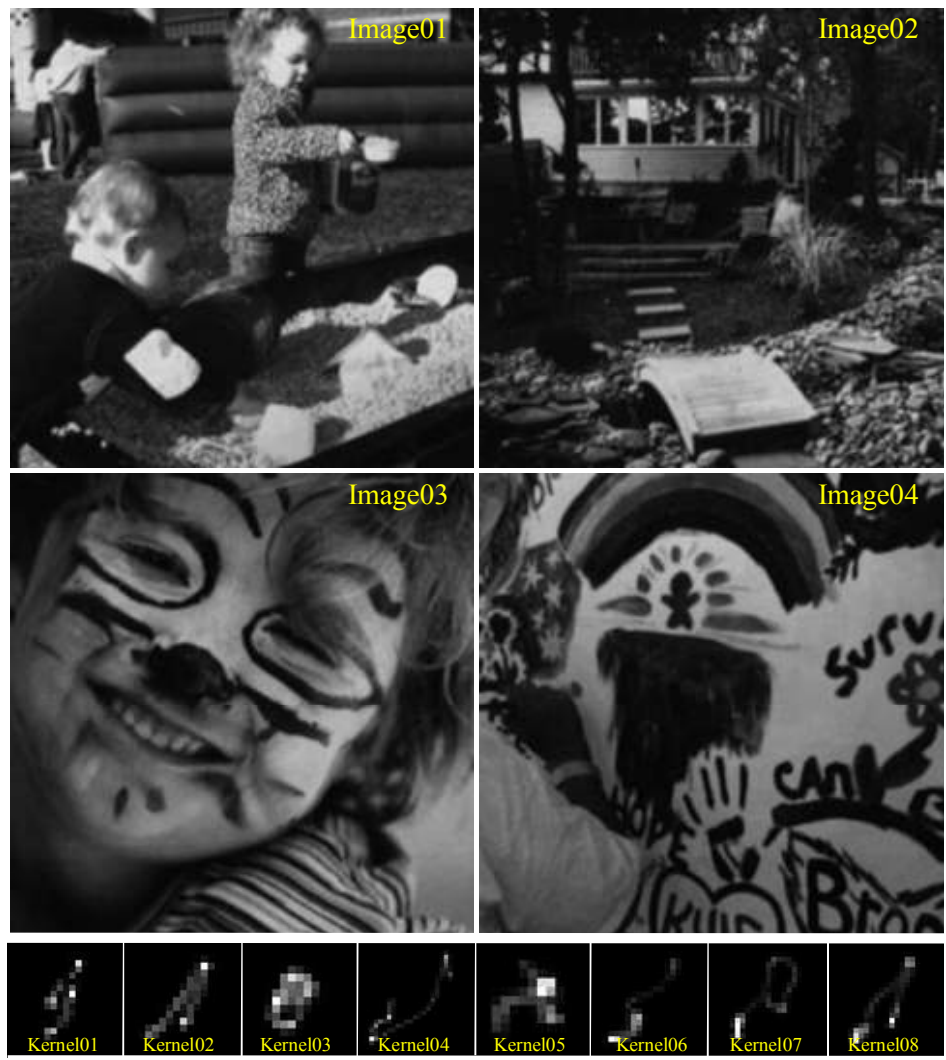

Figure 2. The ground truth images and motion blur-kernels from the benchmark image dataset proposed by Levin *et al.* [2].

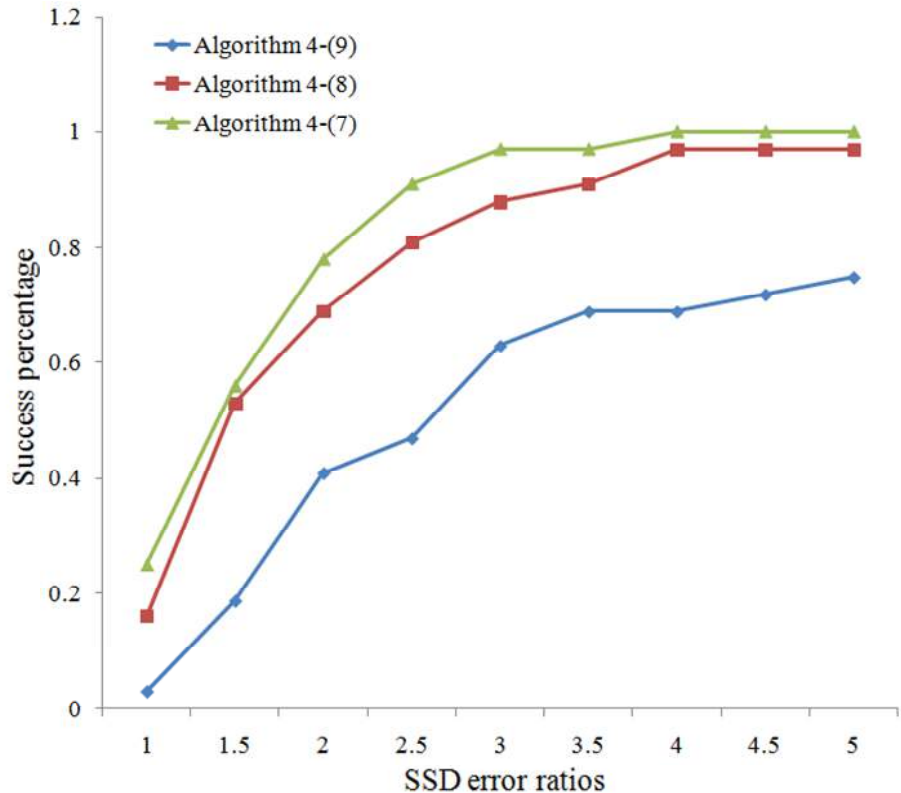

Figure 3. The cumulative histogram of the SSD deblurring error ratios achieved by **Algorithm 4** utilizing different regularization constraints (7)-(9) introduced in Section 2. For each bin, the higher the bar, the better the blind motion deblurring performance. The proposed method, i.e., **Algorithm 4**-(7), takes the lead with 97% of SSD error ratios below 3.

The cumulative histogram in Figure 3 shows the high success percentage of the proposed method – 97% for **Algorithm 4**-(7); its average SSD error ratio is 1.56, as shown in Table 2. As for **Algorithm 4**-(8) and **Algorithm 4**-(9), their percentages of success are 88% and 63%, and their average SSD error ratios are correspondingly 1.80 and 3.15. According to the results, the performance of blind motion deblurring has greatly improved when incorporating the $\ell_2$-norm-based image prior and the $\ell_0$-norm-based kernel prior into Equation (9), hence convincing the rational of the proposed bi-$\ell_0$-$\ell_2$-norm regularization.

For visual perception and considering the limited space, we just show the deblurring results in Figure 4 (including the estimated blur-kernel, the intermediate sharp image, and the final deconvolution image) produced by each approach for the motion blurred image Image04-kernel06, which is the only failure case (its SSD error ratio is above 3) of the proposed approach, i.e. **Algorithm 4**-(7). The peak signal-to-noise ratio (PSNR) metric is utilized to quantitatively measure the deblurring performance of different algorithms. We observe that the superiority of **Algorithm 4**-(7) to **Algorithm 4**-(8) and **Algorithm 4**-(9) is also shown fairly well in this failure case. Particularly, the intermediate sharp image produced by **Algorithm 4**-(7) has less staircase artifacts than those by its two degenerated versions, naturally leading to more accurate blur-kernel and better final deconvolution image.

Motion blur-kernels     Intermediate sharp images    Final deconvolution images

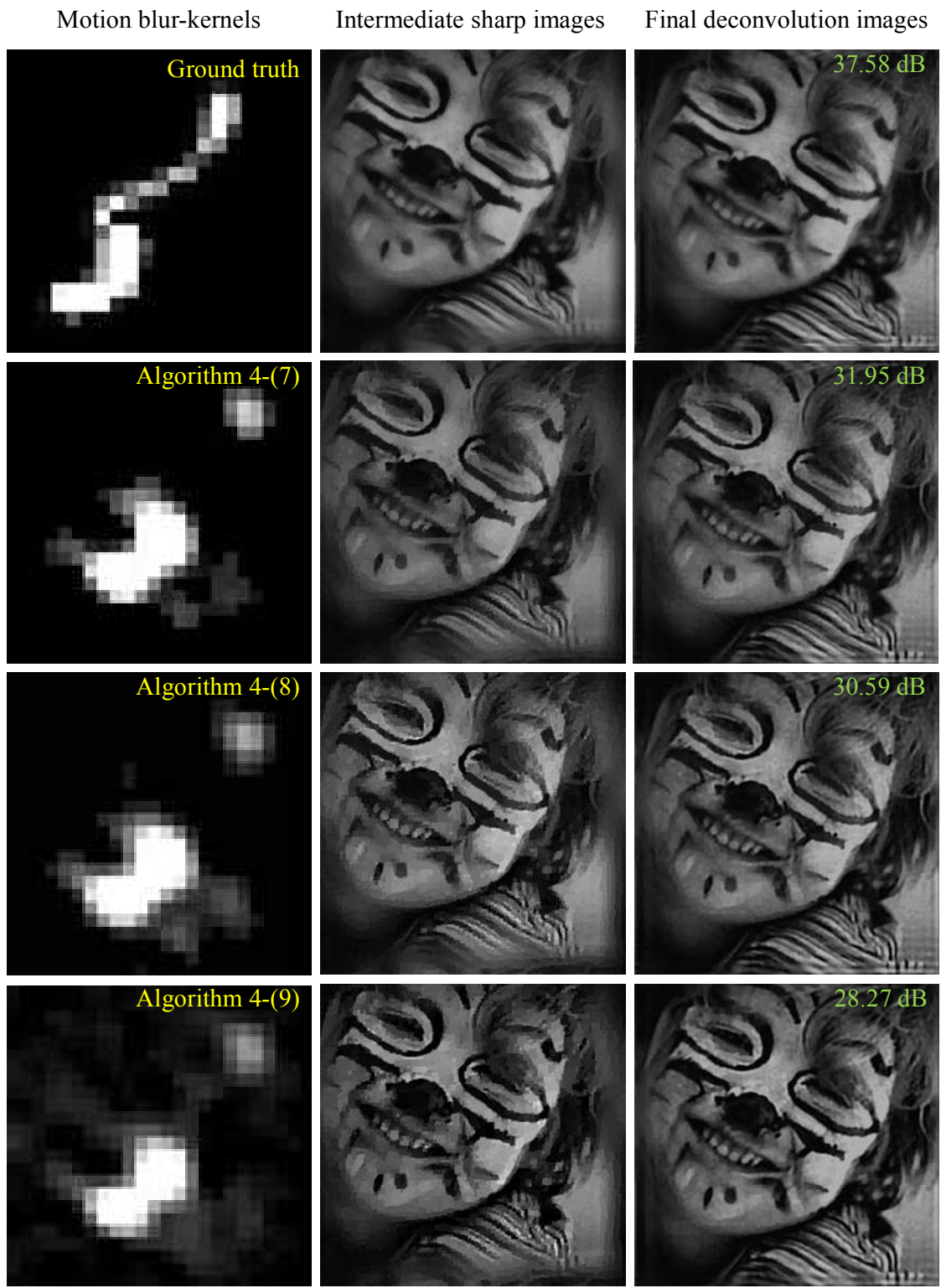

Figure 4. Deblurring results produced by **Algorithm 4**-(7), **Algorithm 4**-(8), and **Algorithm 4**-(9) for the motion blurred image Image04-kernel06, which is the only failure case (its SSD error ratio is above 3) of the proposed method, i.e., **Algorithm 4**-(7). Left: blur-kernels (gray-scale transformed and 5 times interpolated); Middle: intermediate sharp images; Right: final deconvolution images. See the intermediate sharp images and motion blur-kernels on a computer screen for better visual perception.

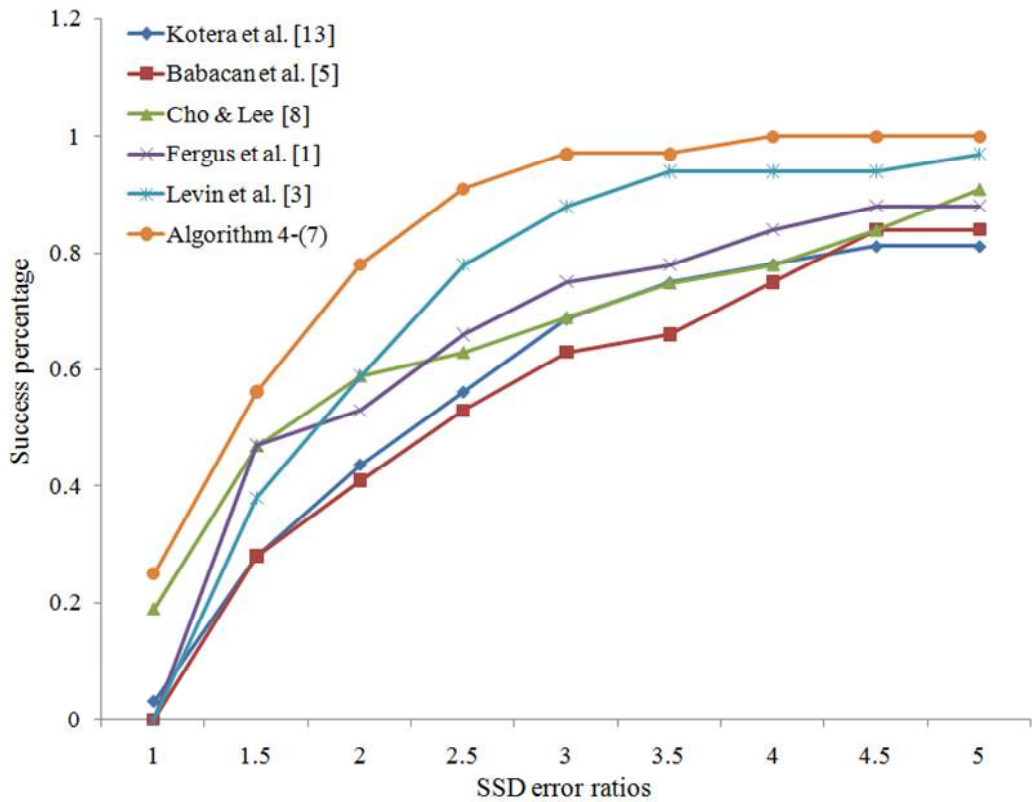

Figure 5. The cumulative histogram of the SSD error ratios achieved by Fergus *et al.*[1], Levin *et al.* [3], Babacan *et al.*[5], Cho & Lee [8], Kotera *et al.* [13], and Proposed, i.e., **Algorithm 4**-(7). The success percentages, i.e., SSD error ratios below 3, of different methods are: 75% [1], 88% [3], 63% [5], 69% [8], 63% [13], 97% (Proposed).

Table 2. Average SSD error ratios and percentages of success achieved by
the proposed approach and other compared methods

| Method | Average SSD Error Ratio | Percentage of Success |
|---|---|---|
| **Algorithm 4**-(7) | **1.56** | **97%** |
| **Algorithm 4**-(8) | 1.81 | 88% |
| **Algorithm 4**-(9) | 3.15 | 63% |
| Fergus *et al.*[1] | 13.5 | 75% |
| Levin *et al.*[3] | 2.06 | 88% |
| Babacan *et al.* [5] | 2.94 | 63% |
| Cho & Lee [8] | 2.67 | 69% |
| Kotera *et al.* [13] | 2.77 | 69% |

In the next group of experiments, the proposed method is compared with the three methods accompanying the benchmark image dataset including Fergus *et al.* [1], Levin *et al.* [3], and Cho & Lee [8], as well as other two recent methods, i.e., Babacan *et al.* [5] and Kotera *et al.* [13]. To be noted that, in the benchmark dataset the SSD deconvolution error ratios of [1], [3], [8] are calculated using the deconvolution images generated by the non-blind deblurring algorithm [28]. As for [5] and [13], motion blur-kernels are estimated by running the provided MATLAB codes by the authors, while the final deconvolution images are obtained using the fast non-blind deblurring algorithm [16], just the same as our proposed approach (including the parameter settings). Figure 5 shows the cumulative histogram of SSD error ratios for the compared five methods [1], [3], [5], [8], [13] as well as **Algorithm 4**-(7).

The percentages of success, i.e., SSD error ratios below 3, of the five compared methods are: 75% [1], 88% [3], 63% [5], 69% [8], and 69% [13]. Their achieved average SSD error ratios are also provided respectively in Table 2. It is seen that the proposed approach (**Algorithm 4**-(7)) achieves the best performance in both terms of average SSD error ratio and success percentage. Also evident from Figure 5 is that our method achieves uniformly good performance throughout all bins. Interestingly, the average SSD error ratio of the VB method [1] is much worse compared to others but with a relatively higher percentage of success. The reason is that there are few examples in the benchmark image dataset for which the VB method [1] fails drastically (more details in [2]).

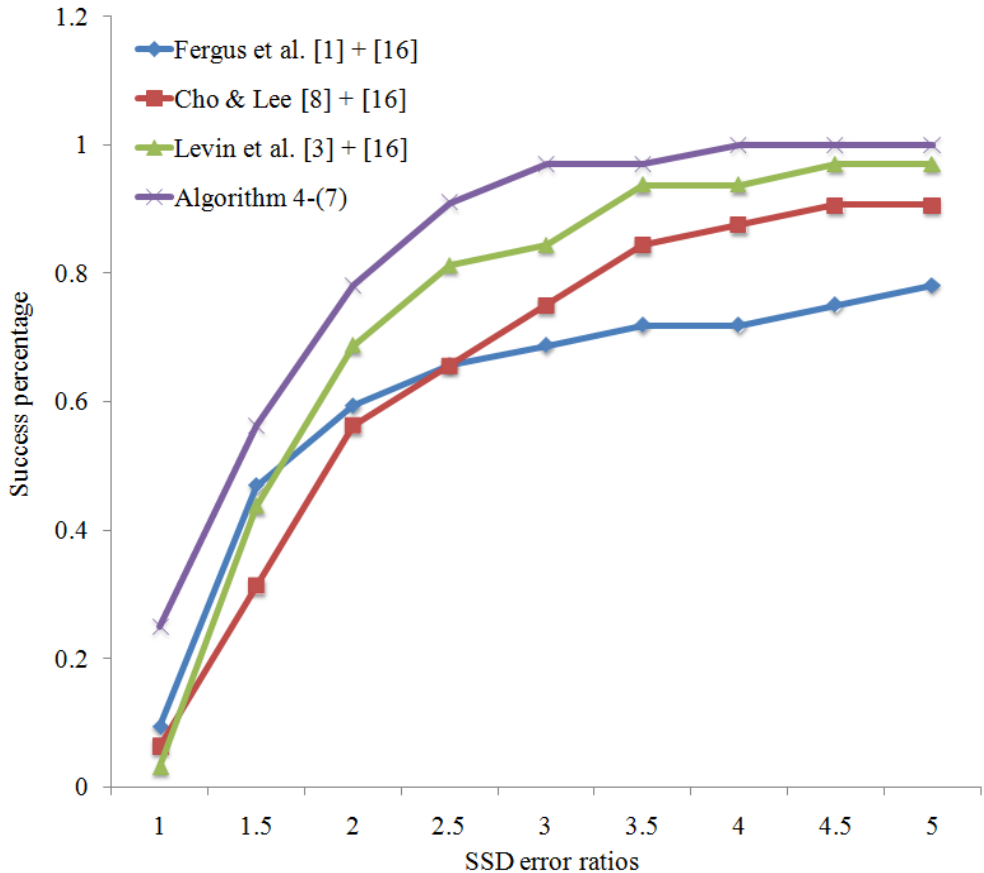

Figure 6. The cumulative histogram of the SSD error ratios achieved by Fergus *et al.*[1], Levin *et al.* [3], Cho & Lee [8], and Proposed, i.e., **Algorithm 4**-(7), using the same final non-blind image deblurring algorithm [16] (including the parameter settings). The success percentages, i.e., SSD error ratio below 3, of different approaches in this case are: 69% [1], 84% [3], 75% [8], 97% (Proposed).

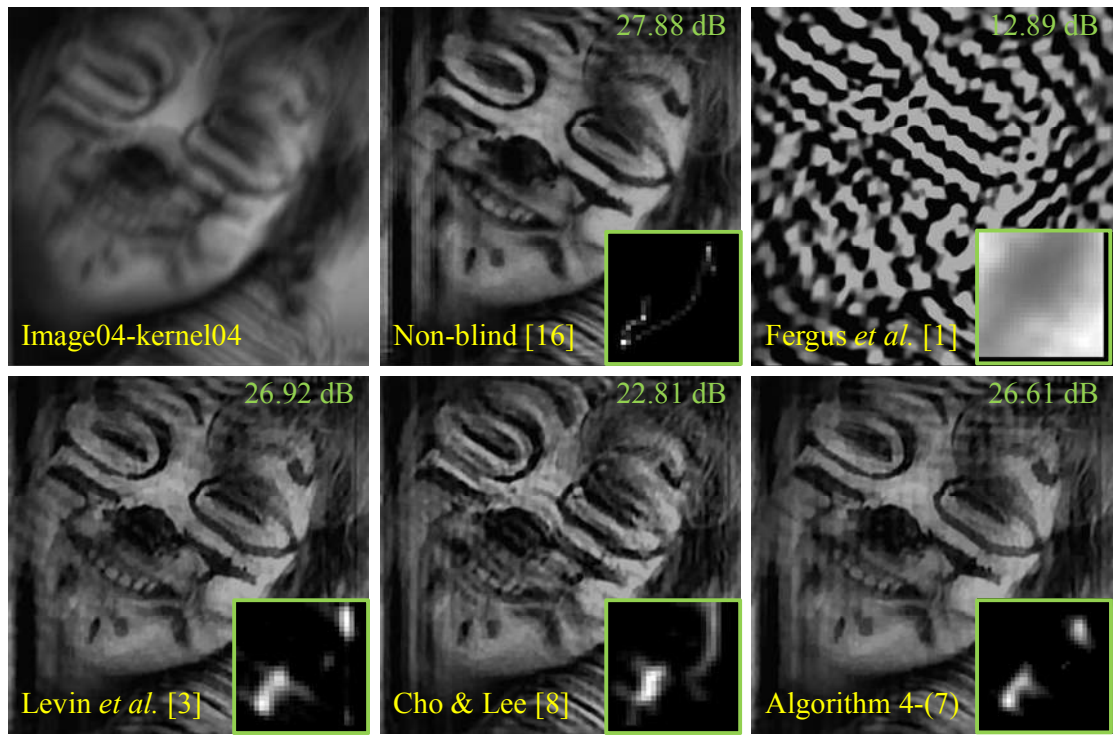

Figure 7. Blind motion deblurring for Image04-kernel04 in the benchmark image dataset [2]. Left to right, top to bottom: motion burred image, non-blind deblurring [16], blind deblurring using Fergus *et al.* [1], Levin *et al.* [3], Cho & Lee [8], and **Algorithm 4**-(7).

One more issue to be discussed is the influence of the final non-blind deblurring method on the SSD error ratios and its influence on the comparison among different methods. We take methods [1], [3], [8] for example, and in the following, the deblurred images corresponding to these methods are generated utilizing the non-blind deblurring algorithm in [16] rather than [28], the same as our method including the parameter settings. In this case, the average SSD error ratios of various methods are now 3.74 for Fergus *et al.* [1], 2.02 for Levin *et al.* [3], 2.42 for Cho & Lee [8]. Comparing with those shown in Table 2, the non-blind deblurring method [16] leads to an improvement of the average SSD error ratio for all the three methods, and particularly as for [1], meaning that [16] is more appropriate than [28] in generating higher quality final deblurred images. With the above changes, the success percentages of the three methods are now[8] 69% for Fergus *et al.* [1], 84% for Levin *et al.* [3], and 75% for Cho & Lee [8]. Still, our approach outperforms the other three methods. In Figure 6, the cumulative histogram of SSD error ratios is shown for each method. It is seen that the proposed method achieves higher success percentage than the other methods in each bin. Therefore, we believe that future comparisons among different motion blur-kernel estimation approaches should be made based on the same non-blind deblurring algorithm. However, many current methods do not follow this rationale, e.g. [4]-[7], [9], [11]-[15]. For visual perception of the final deblurred image corresponding to each motion blur-kernel estimation method, the deblurred images as well as the motion blur-kernels are shown in Figure 7. Here, due to limited space, we only take Image04-kernel04 for example. It is clearly observed that the deblurred image of our method is of better visual perception than the other methods (in spite that its PSNR is slightly lower than that of Levin *et al.* [3]), in particular compared with those of Fergus *et al.* [1] and Cho & Lee [8].

[8] In terms of this percentage measure, not all methods have improved. Nevertheless, the more important quality measure of average SSD error ratio does show the stated improvement.

Figure 8 presents plots of the functionals (10) for updating the sharp image and (11) for updating the motion blur-kernel, in order to demonstrate the convergence tendency of the proposed algorithm. We just refer to the experiment with Image04-kernel04 as a representative example. The graphs show the energy curves of 10 outer iterations for each scale of **Algorithm 4-(7)**. From these curves we see that the proposed OSAL-based alternating minimization algorithm is quite effective in pursuing the (possibly local) minimizers of the functionals (10) and (11).

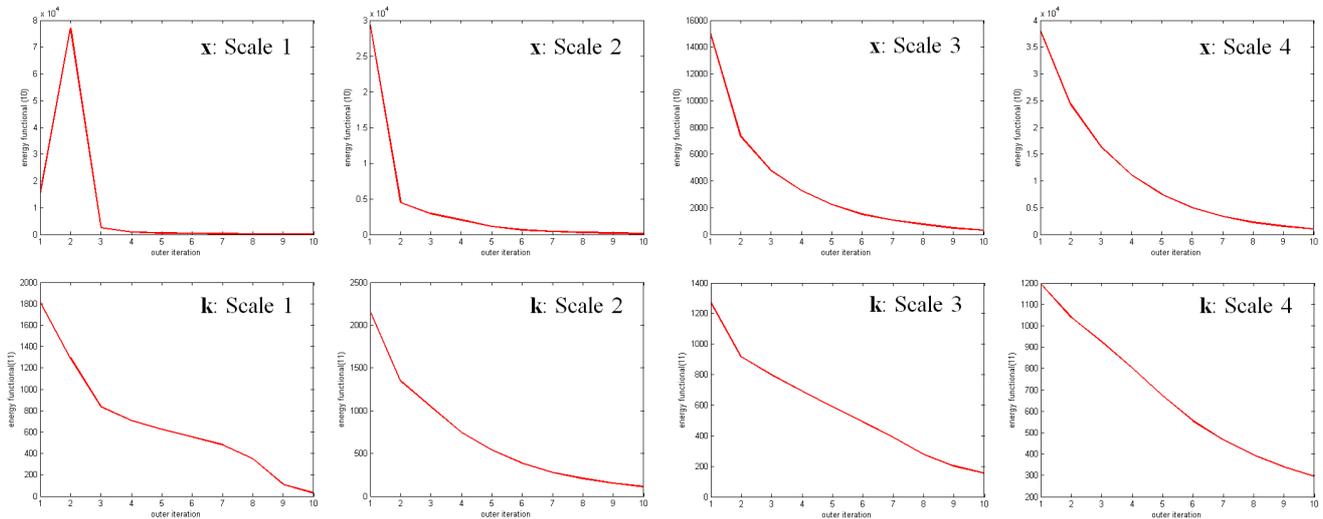

Figure 8. Energy curves of 10 outer iterations for each scale of **Algorithm 4**-(7) as for Image04-kernel04. Top row: functional (10) for estimating **x**; Bottom row: functional (11) for estimating **k**.

The next set of experiments aims to compare the proposed approach with Xu *et al.* [14] as well as its improved version [14]+[9]. As analyzed above, for a completely fair comparison, final image deconvolution for all approaches utilizes the same non-blind deblurring algorithm [16] including the parameter settings; that is, the blur-kernel is produced by the code of each kernel estimation method, with which the final deconvolution image is then generated by [16]. In addition, three different settings of the blur-kernel size are considered for a comprehensive comparison among the different approaches: ground truth (**G**); medium scale (**M**), i.e., 31×31 (in the terminology of [14]), and large scale (**L**), i.e., 51×51. Apparently, the latter two scenarios correspond to blind motion deblurring without any accurate size information on the blur-kernel. It is noted that, in general the larger the blur-kernel size, the harder the blind deblurring problem becomes. It also deserves pointing out that all the approaches are free of parameter adjustment and therefore, the comparisons we provide are fair ones.

Table 3. The SSD error ratios of the 32 test images corresponding to distinct settings of the blur-kernel size (**G**-ground truth, **M**-medium scale, **L**-large scale), achieved by the proposed method, Xu *et al.* [14], and its improved version [14] + [9] with the same non-blind deblurring algorithm [16].

| Proposed | Image01 | | | Image02 | | | Image03 | | | Image04 | | |
| --- | --- | --- | --- | --- | --- | --- | --- | --- | --- | --- | --- | --- |
| | G | M | L | G | M | L | G | M | L | G | M | L |
| Kernel01 | 1.02 | 0.99 | 1.86 | 2.97 | 2.44 | 1.52 | 0.99 | 0.94 | 0.89 | 2.16 | 1.99 | 1.88 |
| Kernel02 | 0.99 | 1.01 | 0.96 | 1.58 | 1.25 | 1.40 | 0.98 | 0.92 | 1.00 | 1.93 | 2.06 | 2.08 |
| Kernel03 | 1.20 | 1.23 | 1.31 | 2.06 | 2.07 | 1.93 | 0.92 | 0.99 | 0.86 | 1.55 | 1.76 | 1.46 |
| Kernel04 | 0.85 | 0.85 | 1.04 | 1.28 | 1.28 | 2.43 | 1.25 | 1.25 | 1.88 | 1.32 | 1.32 | **4.76** |
| Kernel05 | 1.07 | 1.30 | 1.11 | 2.08 | 1.81 | 1.76 | 1.22 | 1.27 | 1.28 | 1.92 | 2.36 | 2.20 |
| Kernel06 | 1.97 | 1.71 | 2.69 | 1.93 | 1.96 | **3.69** | 1.55 | 1.16 | 2.11 | **3.66** | **3.60** | **3.74** |
| Kernel07 | 1.19 | 1.12 | 1.59 | 2.38 | 2.67 | 2.40 | 1.43 | 1.55 | 1.72 | 2.53 | 2.54 | 3.00 |
| Kernel08 | 0.72 | 0.74 | 0.67 | 1.46 | 1.58 | 1.69 | 0.90 | 0.86 | 0.96 | 0.90 | 0.86 | 0.91 |
| **[14] + [9]** | Image01 | | | Image02 | | | Image03 | | | Image04 | | |
| | G | M | L | G | M | L | G | M | L | G | M | L |
| Kernel01 | 2.31 | 2.23 | 2.42 | 2.80 | 2.11 | **3.43** | 1.55 | 1.47 | 1.46 | 2.31 | 2.38 | **4.25** |
| Kernel02 | 1.71 | 1.67 | 1.90 | 2.23 | 2.55 | **3.98** | 1.26 | 1.32 | 1.31 | 1.93 | 2.09 | **4.59** |
| Kernel03 | 2.49 | 2.69 | 2.89 | 1.52 | 2.07 | **3.06** | 2.02 | 1.68 | 1.81 | **3.06** | 2.06 | **3.14** |
| Kernel04 | 1.11 | 1.07 | 1.07 | 1.20 | 1.32 | 1.58 | 1.18 | 1.19 | 1.20 | 1.23 | 1.60 | 1.87 |
| Kernel05 | **3.49** | 2.78 | 2.81 | 2.14 | 1.69 | 1.95 | 2.17 | 1.74 | 1.71 | 2.98 | 2.36 | **3.66** |
| Kernel06 | **5.89** | **4.26** | **4.30** | 2.51 | 2.36 | **3.10** | **4.65** | 2.96 | 2.94 | **6.80** | **3.62** | **4.93** |
| Kernel07 | 1.58 | 1.49 | 1.47 | 1.52 | 1.68 | 1.82 | 1.48 | 1.54 | 1.42 | 2.84 | **3.38** | 25.1 |
| Kernel08 | 0.96 | 0.96 | 0.94 | 1.31 | 1.29 | 1.35 | 1.06 | 0.99 | 1.06 | **6.40** | 0.84 | 0.84 |
| **[14]** | Image01 | | | Image02 | | | Image03 | | | Image04 | | |
| | G | M | L | G | M | L | G | M | L | G | M | L |
| Kernel01 | **3.14** | 2.63 | 2.64 | **3.73** | **3.82** | **6.10** | 1.72 | 1.66 | 1.70 | 2.63 | 2.90 | **5.37** |
| Kernel02 | 1.87 | 1.00 | 2.12 | 2.68 | **5.49** | **5.69** | 1.56 | 1.88 | 1.98 | 1.99 | 2.08 | **5.22** |
| Kernel03 | 2.76 | **3.79** | **3.62** | 2.64 | **3.14** | **4.01** | 1.99 | 1.42 | 1.24 | 1.82 | 2.15 | **5.40** |
| Kernel04 | 1.28 | 1.16 | 1.31 | 1.71 | 1.79 | 2.56 | 1.83 | 1.88 | 1.98 | **5.61** | 1.54 | 2.60 |
| Kernel05 | **5.24** | **4.57** | **3.79** | 3.04 | 2.55 | 2.84 | **3.49** | 1.94 | 1.81 | **3.12** | 2.83 | **6.63** |
| Kernel06 | **8.50** | **5.06** | **5.66** | **6.14** | **4.11** | **5.34** | **6.26** | **3.51** | **3.51** | **5.96** | **4.27** | 6.45 |
| Kernel07 | 2.12 | 1.91 | 1.83 | 2.18 | 1.92 | 2.31 | 1.98 | 1.85 | 1.70 | **3.63** | **4.55** | 29.7 |
| Kernel08 | 1.14 | 1.00 | 1.05 | 1.62 | 1.57 | 2.12 | 1.18 | 1.12 | 1.17 | **6.53** | 0.94 | 2.41 |

Table 3 provides the SSD error ratios of the 32 test images corresponding to different settings of the blur-kernel size, achieved by the proposed approach, Xu *et al.* [14], and its improved version [14]+[9] with the same non-blind image deconvolution algorithm [16]. The percentage of success and the average SSD error ratio are provided for each scenario in Table 4. According to the results, it is obvious that the proposed approach has achieved fairly more robust and precise blur-kernel estimation than either [14] or its extension [14]+[9]. Particularly, the percentage of success and the average SSD error ratio of the proposed method in the case of medium scale kernel size (97%, 1.55) are nearly the same as those in the case of true kernel size (97%, 1.56). With the

Table 4. Percentages of success and average SSD error ratios achieved by the proposed method, Xu *et al.* [14], and its improved version [14]+[9] corresponding to different settings of the blur-kernel size (**G**-ground truth, **M**-medium scale, **L**-large scale)[9].

| Settings | Percentage of Success | | | Average SSD Error Ratio | | |
|---|---|---|---|---|---|---|
| | Proposed | [14]+[9] | [14] | Proposed | [14]+[9] | [14] |
| **G** | **97%** | 81% | 59% | **1.56** | 2.43 | 3.16 |
| **M** | **97%** | 91% | 69% | **1.55** | 1.98 | 2.56 |
| **L** | **91%** | 66% | 56% | **1.83** | 3.11 | 4.21 |

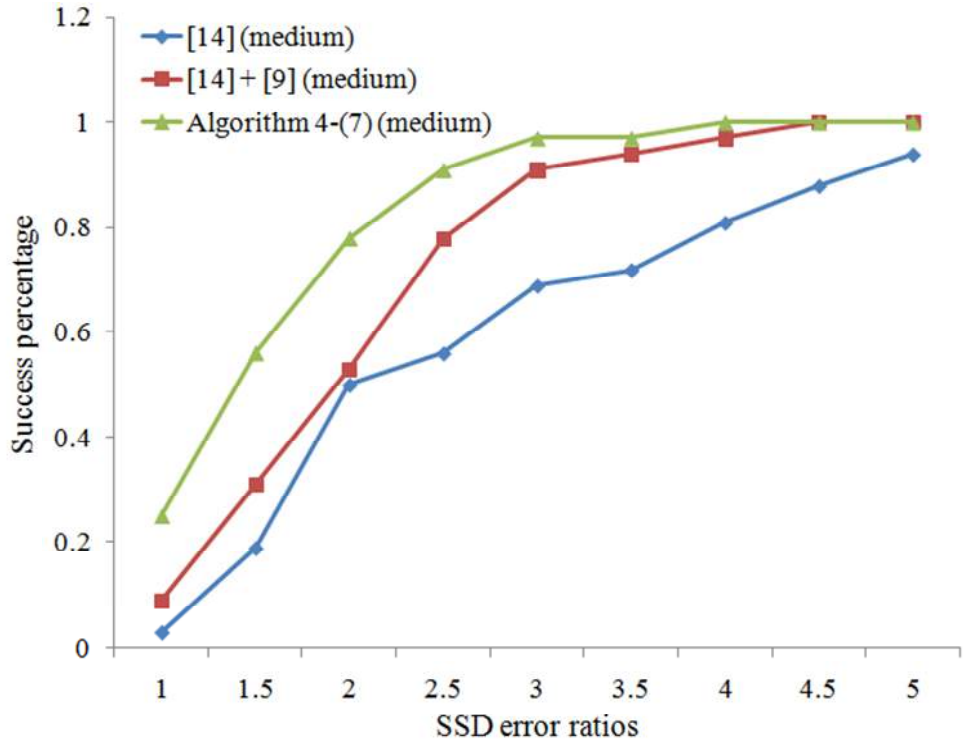

Figure 9. The cumulative histograms of SSD error ratios as the kernel size is of medium scale, i.e., 31×31, achieved by [14], [14]+[9], and the proposed approach, i.e., **Algorithm 4**-(7), using the same final image deconvolution algorithm [16]. Their success percentages, i.e., SSD error ratios below 3, are respectively 69% [14], 91% [14]+[9], 97% (Proposed).

kernel size increasing, it is observed that the average SSD error ratio of the proposed method also increases (1.83), leading to a slightly lower percentage of success (91%). This is natural because a larger blur-kernel implies a more difficult kernel estimation problem with solutions in a higher-dimensional space. In contrast, [14] and its extension [14]+[9] achieve the best performance in the case of medium scale kernel size, i.e., [14] (**M**: 69%, 2.56), [14]+[9] (**M**: 91%, 1.98). However, their performance degrades dramatically in either the case of true kernel size ([14] (**G**: 59%, 3.16), [14]+[9] (**G**: 81%, 2.43)) or large kernel size ([14] (**L**: 56%,

---

[9] We also provide the results obtained with the degenerate **Algorithm 4**-(8) in each setting of the blur-kernel size. They are directly provided here just for readers' reference: ground truth (88%, 1.81); medium scale (78%, 1.97), and large scale (72%, 2.65).

4.21), [14]+[9] (**L**: 66%, 3.11)).

Figure 10. Motion blur-kernel estimation in the case of medium scale kernel size for Image02, i.e., 31×31. Left to right: Ground truth kernels, the proposed approach, i.e., **Algorithm 4**-(7), [14]+[9], [14]. Top to bottom: Kernel01~Kernel08.

In Figure 9, the cumulative histograms of SSD error ratios corresponding to the three approaches are plotted for the case of medium scale kernel size. Observe that the proposed approach performs better than the other two throughout all bins in each setting, demonstrating again the robust performance of the proposed framework with the bi-$\ell_0$-$\ell_2$-norm regularization. In Figure 10, we also provide the 8 estimated motion blur-kernels corresponding to the ground truth image Image02 for the case of medium scale kernel size, obtained respectively by the proposed approach, [14]+[9], and [14]. It is observed that the proposed approach achieves more reliable blur-kernel estimation, with more accurate kernel supports and less false motion trajectories as well as isolated points than the blur-kernels by the other two approaches. One more point to be noted is that in the case of medium scale kernel size, the average running-time of the proposed approach over 32 motion blurred images is about 3.8s (MATLAB), while that of Xu *et al.* [14] is about 1.2s, and [14]+[9] is about 1.4s (C++). All the experiments are performed on the same laptop computer (Dell Latitude E6540), with an Intel i7-4600M CPU (2.90GHz) and 8GB memory, running Windows 7 (Professional, Service Pack 1).

*4.2 Experiments on real-world color motion blurred images*

We conclude this section by testing the proposed approach on several real-world color motion blurred images and comparing it with four previously mentioned methods: a VB method [3] (Levin *et al.*) and three MAP methods including [12] (Cai *et al.*), [13] (Kotera *et al.*), and [14]+[9] (Xu *et al.*). The motion blur-kernel for each approach is generated utilizing the codes and parameter settings provided and suggested by the authors. In each deblurring experiment, the blur-kernel size set for [3], [13], [14]+[9] and the proposed approach is the same, either 19×19 (small scale), 31×31 (medium scale), or 51×51 (large scale). As for [12], the executable software returns the blur-kernel size 65×65 by default and users do not have the freedom of altering this size. Again, the non-blind deblurring algorithm [16] is used to produce the final deconvolution image for each kernel estimation method. Another point to be noted is that, in our method the blur-kernel is estimated using the gray version of the color image.

In the first group of experiments, three real-world color motion blurred images, i.e., `Board`, `Fish`, `Roma`, with different sizes and blurring levels are used to test the performance of the above-mentioned five methods. The deblurred images and estimated motion blur-kernels are respectively shown in Figure 11, Figure 12 and Figure 13. It is observed that our proposed approach and the VB method [3] achieve visually plausible blind motion deblurring across the three experiments, regardless of the blurring level, be it large or small. As for other three methods, particularly [12] and [13], their deblurring performance is not as uniform as the proposed approach. Specifically, in Figure 11, notable ringing artifacts in the deblurred image by [14]+[9] are clearly seen, while the other four methods achieve better deblurring performance; in Figure 12, the proposed method generate reasonable motion blur-kernels as well as visually acceptable deblurred images, those of [3] and [14]+[9] are of relatively lower quality, and [13] especially [12] have failed to some extent; in Figure 13, the blind deblurring performance of [3] and [14]+[9] is a bit better than the proposed method, and one might argue that the approaches [12], [13] are completely failing in this example. The evident differences among deblurred images by various methods can be observed in the marked red and yellow circles. The running-time for each approach and each experiment is also provided in Table 5. We see that [14]+[9] (Xu *et al.*) is the most efficient among the five methods. Our method performs more efficiently than the remaining three methods, particularly compared against [3] (Levin *et al.*) and [12] (Cai *et al.*). It is observed that [12] (Cai *et al.*) is of the highest computational complexity among the five methods compared.

Table 5. Running-time (in seconds) of the state-of-the-art methods [3], [12], [13], [14] and the proposed approach for each real-world color motion blurred image

| Image | Image Size | [3] (MATLAB) | [12] (MATLAB) | [13] (MATLAB) | [14] + [9] (C++) | Ours (MATLAB) |
|-------|-----------|-------------|--------------|--------------|-----------------|---------------|
| Board | 480×640 | 345.5 | 3032 | 26.06 | **2.689** | 14.63 |
| Fish | 558×800 | 1331 | 5423 | 31.84 | **4.620** | 28.10 |
| Roma | 417×593 | 3374 | 4488 | 28.89 | **3.121** | 28.16 |

In Figure 14, two more real-world motion blurred images (`Book` and `Boat`), are tested to make further comparisons between the proposed approach and [14]+[9] (Xu *et al.*). For each blurred image, three sizes are assumed for the motion blur-kernel, i.e., 19×19 (small scale), 31×31 (medium scale), and 51×51 (large scale). The final deconvolved images and the motion blur-kernels estimated by the two approaches are provided in Figure 14 (`Book`) and Figure 15 (`Boat`). From Figure 15, it is seen that the proposed method has achieved better deblurring performance than [14]+[9] (Xu *et al.*) in all the three cases of motion blur-kernels. It is also seen that, to some extent, [14]+[9] (Xu *et al.*) fails to recover the true support of the motion blur-kernel as for both small and medium scale cases; while in these cases the proposed approach still generates quite plausible blur-kernels. In this test both methods perform better under the assumption of a large scale blur-kernel. These experiments demonstrate again that the proposed approach is more robust to the motion blur-kernel size than [14]+[9] (Xu *et al.*). The experimental results shown in Figure 16 also demonstrate the superiority of the proposed method to [14]+[9] (Xu *et al.*). In this example, [14]+[9] (Xu *et al.*) fails in all the three cases of motion blur-kernels to a certain degree. On the contrary, the proposed approach has achieved uniformly good deblurring performance in all the cases; its final deconvolution images are of similar visual perception, with much clearer details than those of [14]+[9] (Xu *et al.*). We should add that the proposed approach has been tried on many other real-world motion blurred images with the parameter settings suggested in this paper, most of which are deblurred with visually plausible perception.

## 5. Conclusion

This paper introduces a relatively simple (model-wise), very effective (quality-wise) and efficient (computation-wise) motion blur-kernel estimation method for blind motion deblurring. The core contribution is the proposal of a new sparse model to improve the precision of motion blur-kernel estimation, i.e., the bi-$\ell_0$-$\ell_2$-norm regularization imposed on both the intermediate sharp image and the motion blur-kernel. The motion blur-kernel estimation is formulated as an alternating estimation of the sharp image and the blur-kernel, each corresponding to an $\ell_0$-$\ell_2$-regularized least squares problem, which in turn is solved by a fast numerical algorithm which couples the operator splitting and the augmented Lagrangian techniques. The blind deblurring performance of the proposed approach is intensively validated via a long series of experiments on both Levin *et al.*'s benchmark image dataset and five real-world color motion blurred images. The experimental results demonstrate that the proposed method is highly competitive with state-of-the-art blind motion deblurring methods in both deblurring effectiveness and computational efficiency.

## Acknowledgements

We would like to show our gratitude to the authors of Refs. [3], [5], [12], [13], [14] for their provided image dataset and software used in this paper. The first author Wen-Ze Shao is grateful to Professor Zhi-Hui Wei, Professor Yi-Zhong Ma and Dr. Min Wu, and Mr. Ya-Tao Zhang for their kind support in the past years. This research was supported by the European Research Council under EU's 7th Framework Program, ERC Grant agreement no. 320649, and by the Intel Collaborative Research Institute for Computational Intelligence, the National Natural Science Foundation (NSF) of China (61402239), the NSF of Government of Jiangsu Province (BK20130868), the NSF for Jiangsu Institutions (13KJB510022), and the NSF of NUPT (NY212014).

## Appendix A - Prior Art.

We bring below a brief description of earlier work on blind motion deblurring, both VB and MAP based. Our discussion here emphasizes the choice of the priors for the kernel and the image, as these are the main theme in this paper.

**VB Methods.** Fergus *et al.* [1] use a mixture-of-Gaussians prior to model the image and a mixture-of-Exponentials prior to model the motion blur-kernel, with hyper-parameters in both priors learned in advance; in their experiments they empirically find that the MAP formulation with the same image and kernel priors completely fails. Inspired by Fergus *et al.*'s original work, Levin *et al.* [2], [3] provide a more profound analysis of the blind motion deblurring problem, based on which a simpler VB posteriori inference scheme is deduced, assuming a non-informative uniform (NIU) prior on the kernel. In [4], Amizic *et al.* impose a hyper-Laplacian prior [16] on the image, and a total-variation prior [17] on the blur-kernel. Recently, a new blind motion deblurring approach has been proposed by Babacan *et al.* [5], imposing a general sparsity-inspired prior on the image using its integral representation (scale mixture of Gaussians) as well as the NIU prior on the blur-kernel. Interestingly, the non-informative Jeffreys (NIJ) image prior has been practically demonstrated to be more powerful than other options, e.g., [1]-[4]. This finding is highly consistent with the theoretical presentation in [6], which suggests that the NIJ prior is optimal to a certain degree. Particularly, in the noiseless case, the optimal posteriori mean estimates of $x$ and $k$ in [5], [6] can be approximately reformulated through the formulation posed in Equation (3) with $\lambda \to +\infty$ and $\mathscr{R}_x(x)$, $\mathscr{R}_k(k)$ defined as $\mathscr{R}_x(x) = \sum_m \log |(\nabla \mathbf{x})_m|$, $\mathscr{R}_k(k) = \log \| \mathbf{k} \|_2$, where $\mathbf{x}$ and $\mathbf{k}$ are the vectorized versions of the image $x$ and the blur-kernel $k$, and $\nabla \triangleq (\nabla_h; \nabla_v)$ with $\nabla_h, \nabla_v$ the first-order difference operators in horizontal and vertical directions. Indeed, it is not hard to see that the above choice of $\mathscr{R}_x(x)$ is actually highly related to the $\ell_0$-norm via the relationships [6], [21]: $\lim_{p \to 0} \sum_m |(\nabla \mathbf{x})_m|^p = \| \nabla \mathbf{x} \|_0$ and $\sum_m \log |(\nabla \mathbf{x})_m| = \lim_{p \to 0} \frac{1}{p} \sum_m (|(\nabla \mathbf{x})_m|^p - 1)$.

**MAP Methods.** Though the VB approach has enjoyed both theoretical and empirical success, the posteriori inference for VB blind motion deblurring with sparse image and motion blur-kernel priors remains a challenging task [3], [5], [6]. Compared to the VB methods, the MAP principle is practiced more commonly and achieves comparative deblurring perfor- mance in general. There are several key advantages to the MAP approach: (i) it is intuitive; (ii) it offers a simple problem formulation; (iii) it is flexible in choosing the regularization terms; and (iv) it typically leads to an efficient numerical implementation. Of course, many approaches of this category also exploit sparse priors, particularly on the image, either explicitly or implicitly. However, as shown in [2], [6], [22], during motion blur-kernel estimation, common natural image statistics (e.g., $\ell_p$-norm-based super-Gaussian ( $1 \geq p \gg 0$ ) [29], Fields-of-Experts [30] and its extension [31]), tend to fail, as they prefer a blurred image over a sharper one. As a result, unnatural sparse image priors are more advocated in the literature.

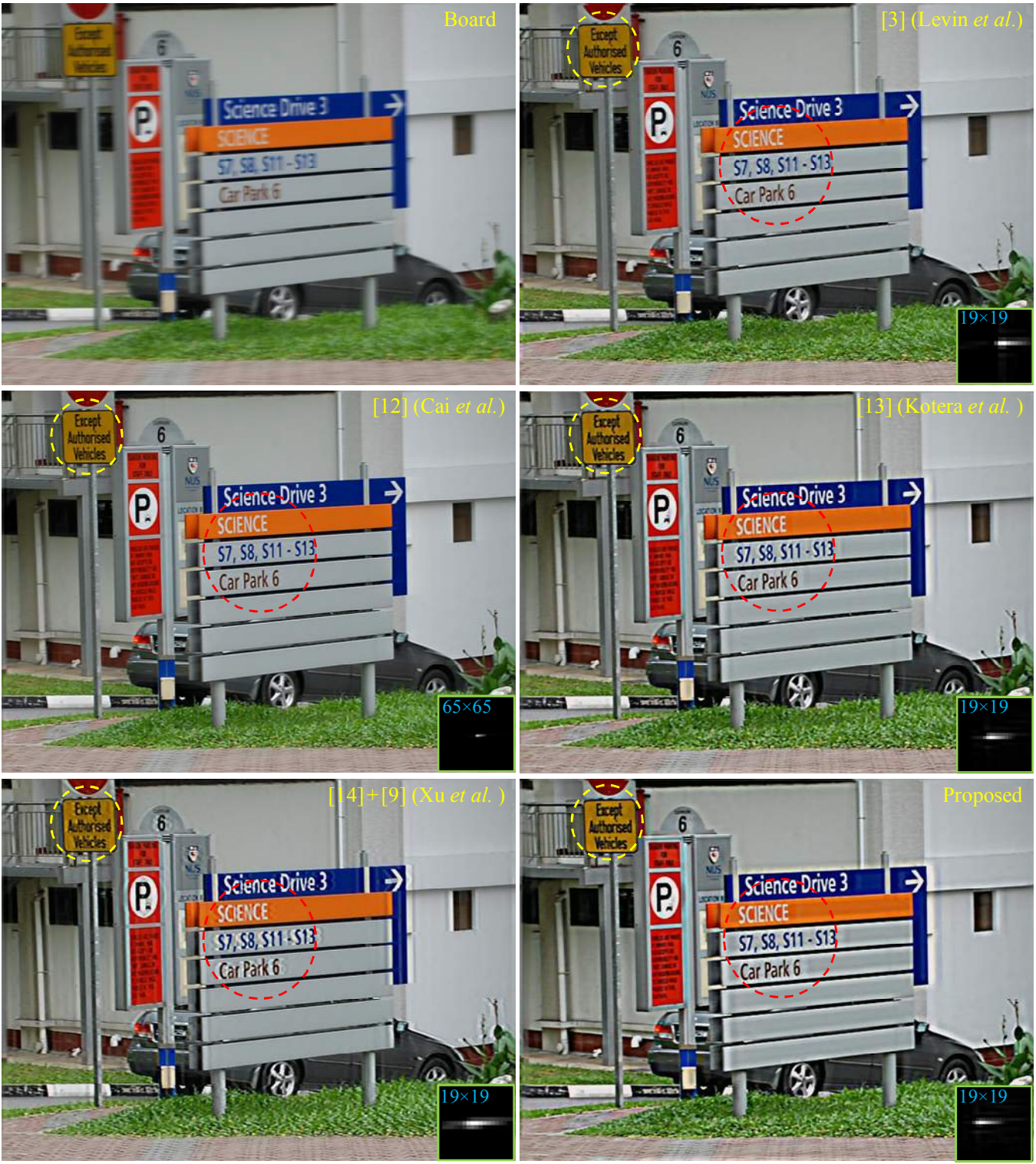

Figure 11. Blind motion deblurring with blurred image `Board`. Left to right, top to bottom: blurred image, deblurred images and estimated kernels by [3] (Levin *et al.*, kernel size: 19×19), [12] (Cai *et al.*, kernel size: 65×65), [13] (Kotera *et al.*, kernel size: 19×19), [14] + [9] (Xu *et al.*, kernel size: 19×19), and the proposed approach (**Algorithm 4**-(7), kernel size: 19×19).

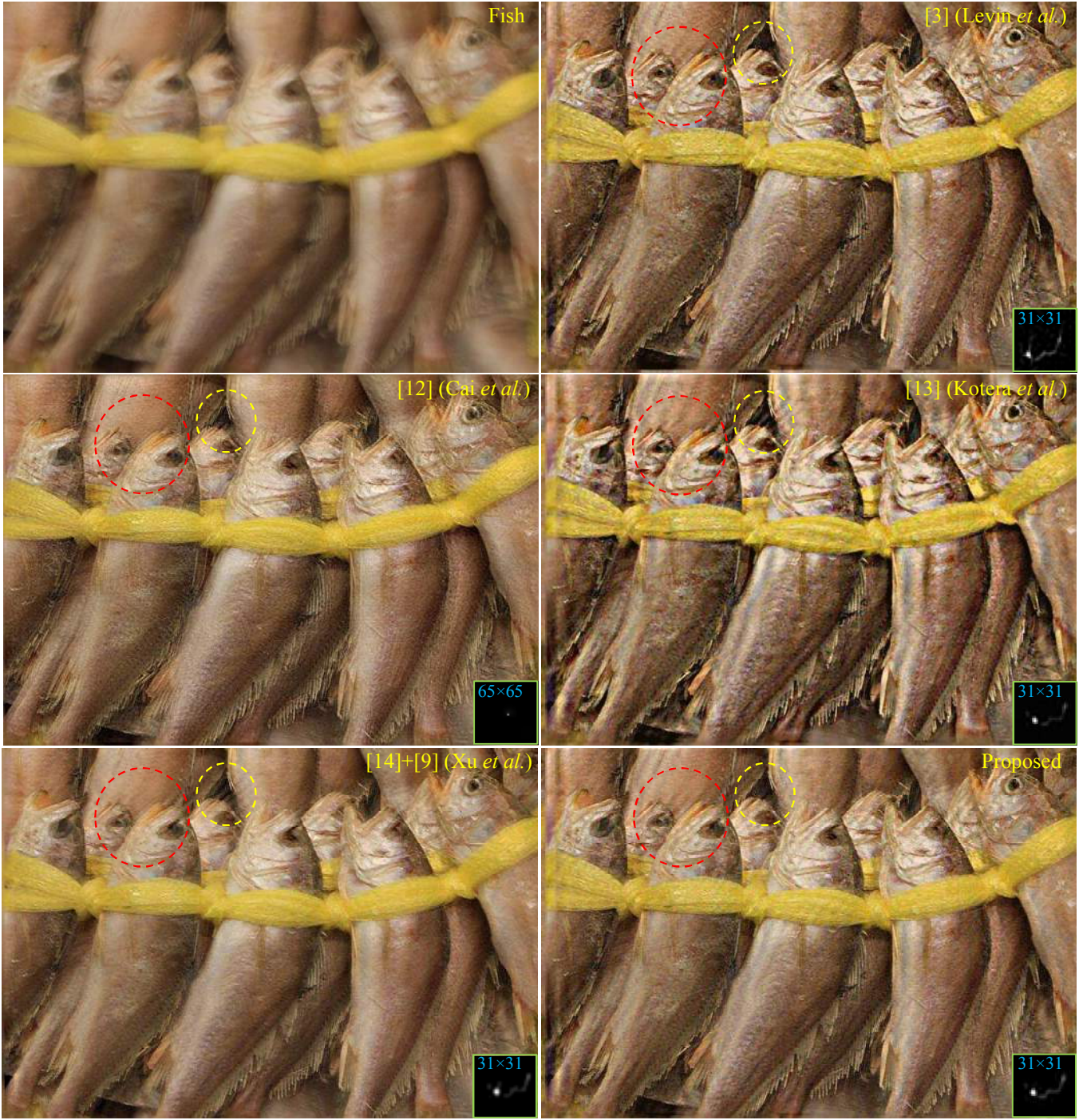

Figure 12. Blind motion deblurring with blurred image `Fish`. Left to right, top to bottom: blurred image, deblurred images and estimated kernels by [3] (Levin *et al.*, kernel size: 31×31), [12] (Cai *et al.*, kernel size: 65×65), [13] (Kotera *et al.*, kernel size: 31×31), [14] + [9] (Xu *et al.*, kernel size: 31×31), and the proposed approach (**Algorithm 4**-(7), kernel size: 31×31).

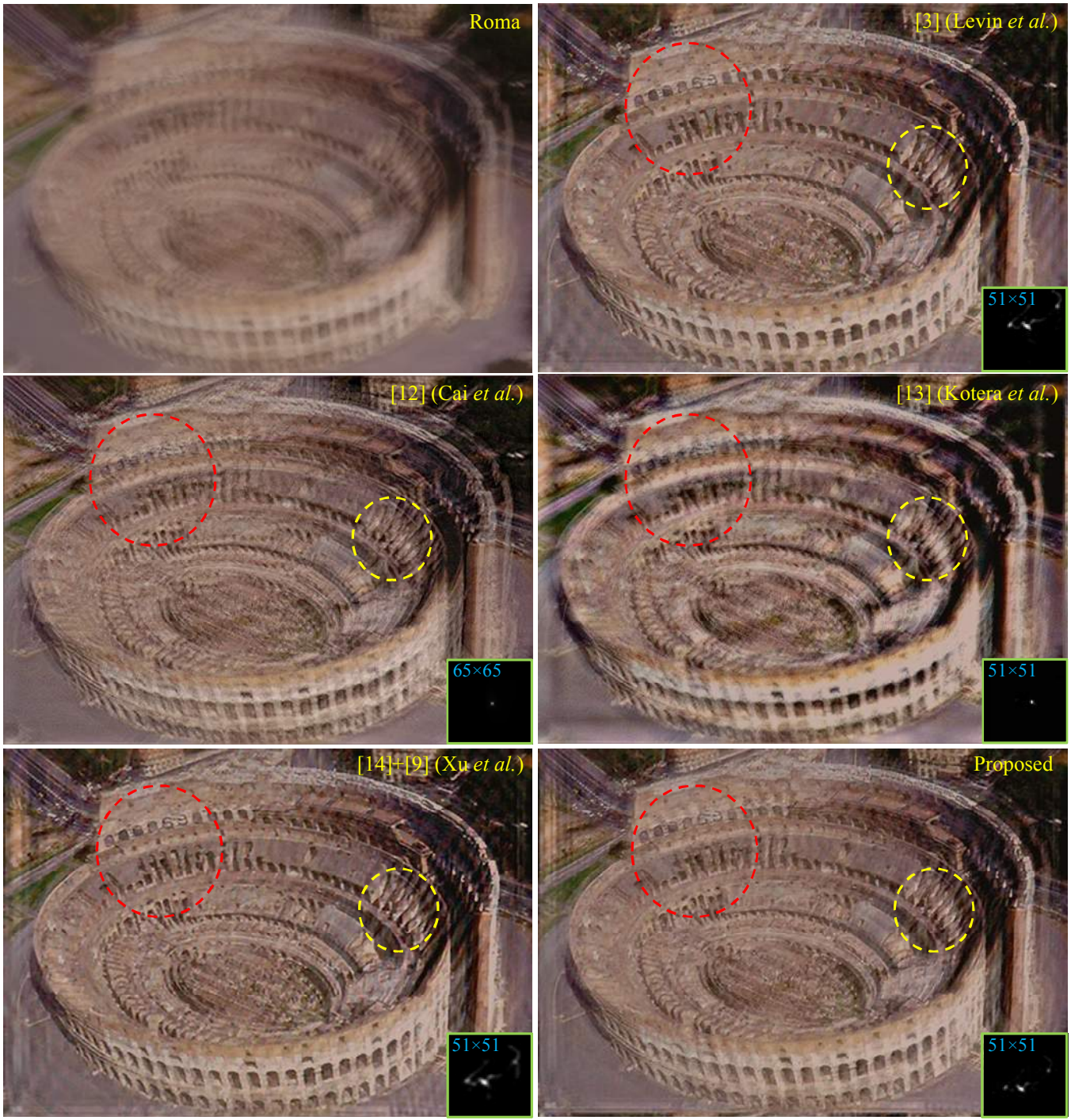

Figure 13. Blind motion deblurring with blurred image Roma. Left to right, top to bottom: blurred image, deblurred images and estimated kernels by [3] (Levin *et al.*, kernel size: 51×51), [12] (Cai *et al.*, kernel size: 65×65), [13] (Kotera *et al.*, kernel size: 51×51), [14] + [9] (Xu *et al.*, kernel size: 51×51), and the proposed approach (**Algorithm 4**-(7), kernel size: 51×51).

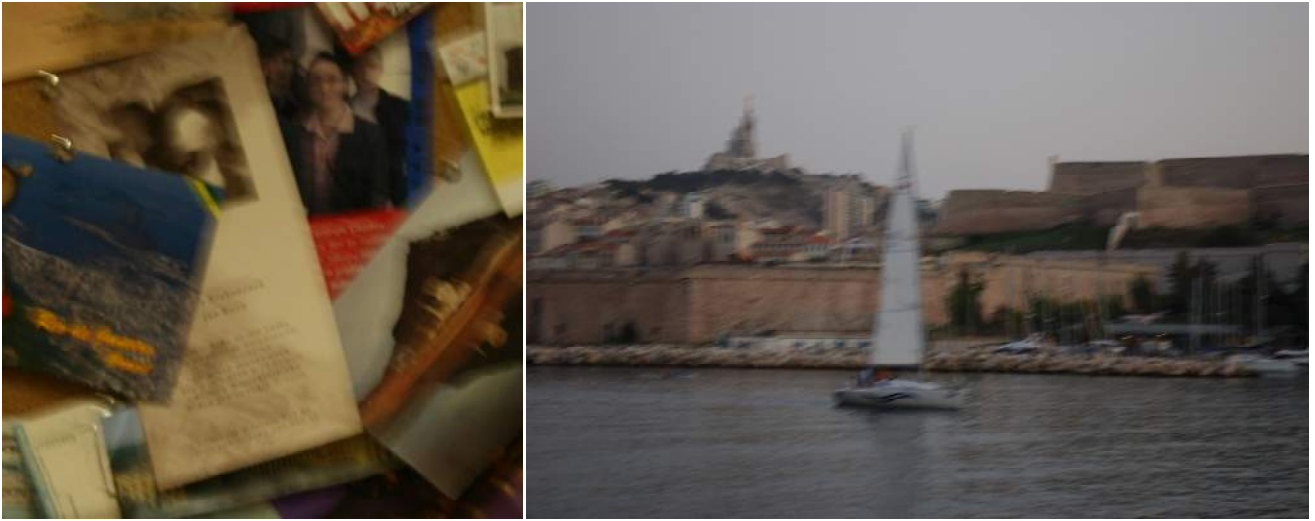

Figure 14. Real-world color motion blurred images Book (800×800) and Boat (1064×1600).

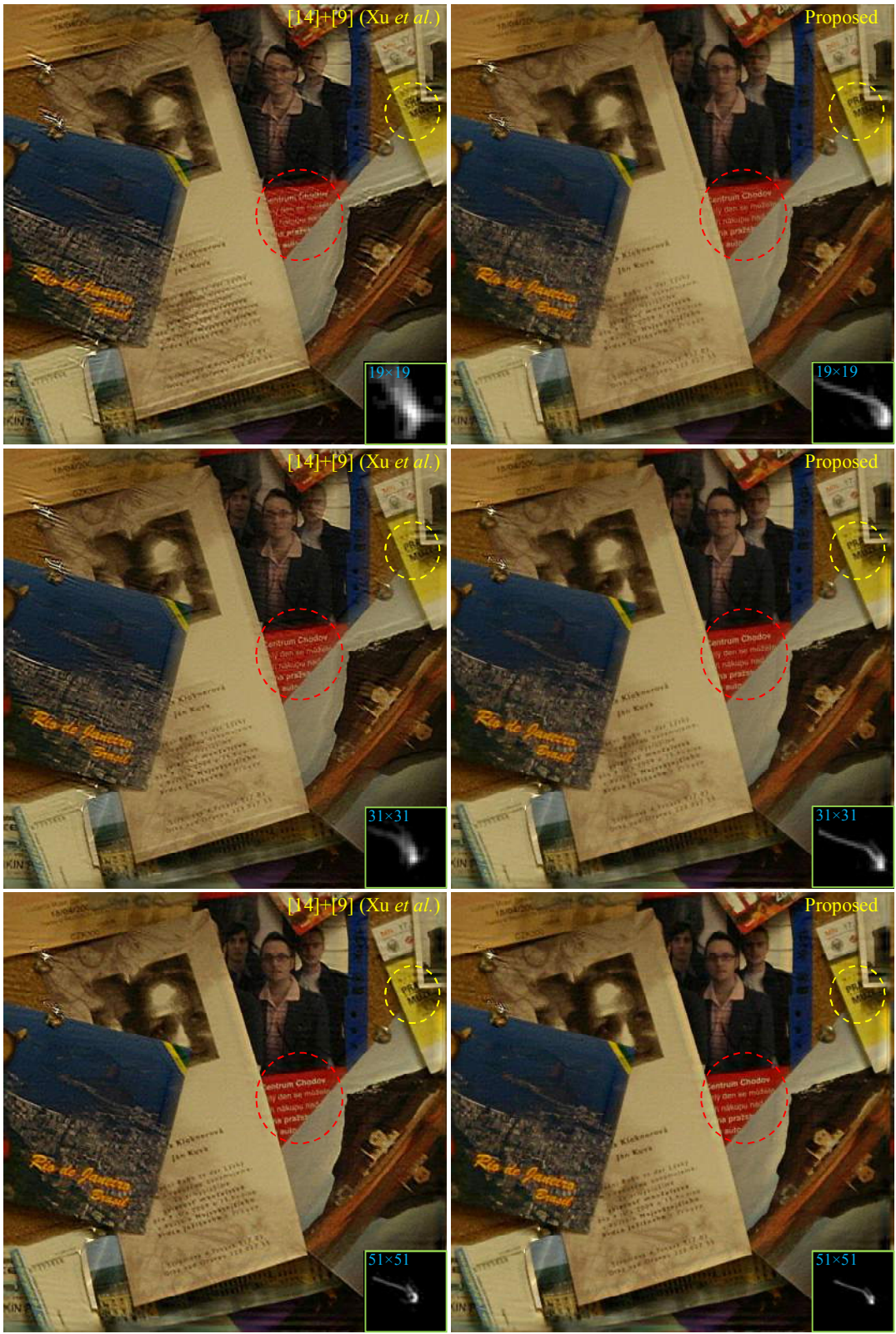

Figure 15. Blind motion deblurring with blurred image `Book`. Left to right: deblurred images and estimated blur-kernels by [14]+[9] ( Xu *et al.*) and the proposed approach (**Algorithm 4**-(7)); Top to bottom: 19×19 (small scale), 31×31 (medium scale), 51×51 (large scale).

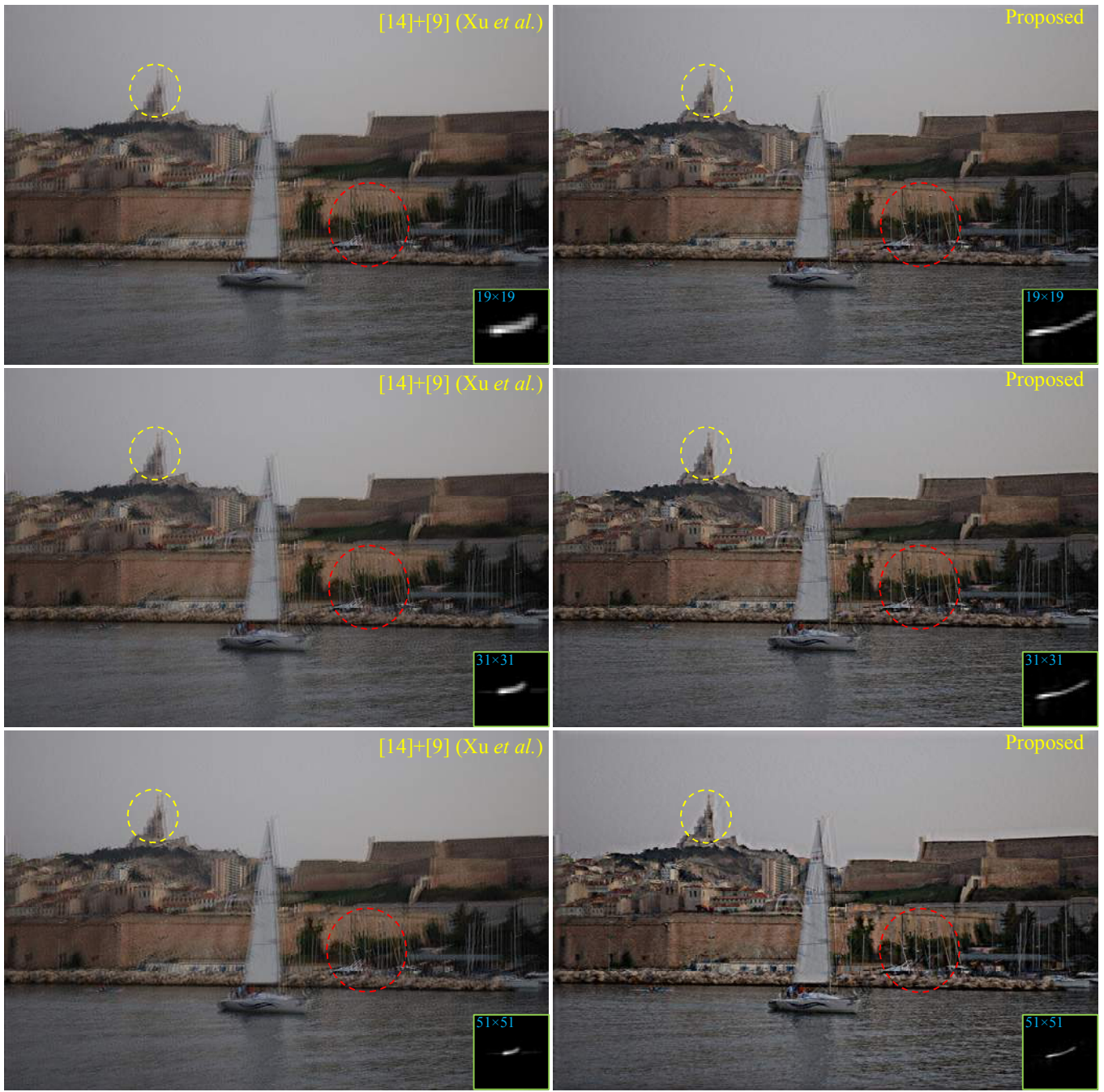

Figure 16. Blind motion deblurring with blurred image `Boat`. Left to right: deblurred images and estimated blur-kernels by [14]+[9] ( Xu *et al.*) and the proposed approach (**Algorithm 4**-(7)); Top to bottom: 19×19 (small scale), 31×31 (medium scale), 51×51 (large scale).